\documentclass{article}

\usepackage{PRIMEarxiv}
\usepackage[utf8]{inputenc} 
\usepackage[T1]{fontenc}    
\usepackage{hyperref}       
\usepackage{url}            
\usepackage{booktabs}       
\usepackage{amsfonts}       
\usepackage{nicefrac}       
\usepackage{microtype}      
\usepackage{graphicx}       
\usepackage{fancyhdr}       
\usepackage{amsmath}        
\usepackage{algorithm}      
\usepackage{algorithmic}    
\usepackage{subcaption}     
\usepackage{siunitx}
\usepackage{multirow}
\usepackage[table,xcdraw]{xcolor}
\usepackage{tikz}

\usepackage{float}
\usepackage{adjustbox}      

\graphicspath{{figures/}}   

\title{GAIA: A Foundation Model for Operational Atmospheric Dynamics
}

\author{
  Ata Akbari Asanjan$^{1}$, Olivia Alexander$^{1}$, Tom Berg$^{2}$, Stephen Peng$^{2}$, Jad Makki$^{2}$, Clara Zhang$^{2}$, Matt Yang$^{2}$,  
  \\
  \textbf{Disha Shidham$^{1}$, Srija Chakraborty$^{1}$, William Bender$^{2}$, Cara Crawford$^{2}$,}
  \\
  \textbf{Arun Ravindran$^{2}$, Olivier Raiman$^{1}$, David Potere$^{2}$, David Bell$^{1}$}
  \\
  $^{1}$Research Institute for Advanced Computer Science (RIACS) at Universities Research Space Association (USRA)\\
  $^{2}$BCG X AI Science Institute\\
  \texttt{\{aakbariasanjan, oalexander, dshidham, schakraborty, dbell\}@usra.edu}\\
  \texttt{\{berg.tom, makki.jad, peng.stephen, zhang.clara, yang.matt,  bender.william,} \\ 
  \texttt{crawford.cara, ravindran.arun, potere.david\normalsize\}@bcg.com} \\
}

\begin{document}
\renewcommand{\L}{\mathcal{L}}
\maketitle


\begin{abstract}
We introduce GAIA (Geospatial Artificial Intelligence for Atmospheres), a hybrid self-supervised geospatial foundation model that fuses Masked Autoencoders (MAE) with self-\textbf{di}stillation with \textbf{no} labels (DINO) to generate semantically rich representations from global geostationary satellite imagery. Pre-trained on 15 years of globally-merged infrared observations (2001-2015), GAIA learns disentangled representations that capture atmospheric dynamics rather than trivial diurnal patterns, as evidenced by distributed principal component structure and temporal coherence analysis. We demonstrate robust reconstruction capabilities across varying data availability (30-95\% masking), achieving superior gap-filling performance on real missing data patterns. When transferred to downstream tasks, GAIA consistently outperforms an MAE-only baseline: improving atmospheric river segmentation (F1: 0.58 vs 0.52), enhancing tropical cyclone detection (storm-level recall: 81\% vs 75\%, early detection: 29\% vs 17\%), and maintaining competitive precipitation estimation performance. Analysis reveals that GAIA's hybrid objectives encourage learning of spatially coherent, object-centric features distributed across multiple principal components rather than concentrated representations focused on reconstruction. This work demonstrates that combining complementary self-supervised objectives yields more transferable representations for diverse atmospheric modeling tasks. Model weights and code are available at: \url{https://huggingface.co/bcg-usra-nasa-gaia/GAIA-v1}.
\end{abstract}



\section{Introduction}

Satellite remote sensing data represents a cornerstone of modern weather monitoring and forecasting systems - particularly the global constellation of GOES (Geostationary Operational Environmental Satellite; National Oceanic and Atmospheric Administration), Himawari (Japan Meteorological Agency), and EUMETSAT (European Organisation for the Exploitation of Meteorological Satellites) providing continuous global observations of atmospheric conditions at high temporal resolution \cite{noaa_gpcp_ir, yang2016satellite}. GOES data and similar data from Himawari and EUMETSAT facilitates numerous mission-critical applications across operational meteorology and climate science: the National Weather Service uses these observations to derive Atmospheric Motion Vectors for numerical weather predictions, NASA's Integrated Multi-satellitE Retrievals for GPM (IMERG) system assimilates GOES measurements to enhance global precipitation retrievals, and operational forecasters employ these data for convective-scale nowcasting and tropical cyclone analysis \cite{huffman2023imerg, forsythe2007atmospheric}. Improved feature extraction from GOES data directly impacts forecast accuracy for extreme weather events, enables earlier warning systems for tropical cyclones, and enhances precipitation estimation in data-sparse regions—critical capabilities for climate adaptation and disaster response \cite{wang2021tropical, yang2022correcting}. 

Despite the transformative potential of GOES data, significant methodological challenges hamper optimal usage. Specifically, representing high-dimensional spatiotemporal dependencies, treating missing or quality-compromised measurements, and extracting scale-invariant features across heterogeneous spatial and temporal domains remain fundamental obstacles \cite{reichstein2019deep, wielicki2013achieving, dubovik2021grand}. These challenges significantly impact the reliability of weather monitoring and forecasting systems, particularly during critical weather events. Current approaches often rely on convolutional autoencoders \cite{romero2015unsupervised} or task-specific deep learning models and struggle to capture the complex interplay between global atmospheric circulation patterns and local weather phenomena \cite{rasp2020weatherbench}.

Recent advances in vision foundation models, particularly Masked Autoencoders (MAE) \cite{he2022masked} and self-\textbf{di}stillation with \textbf{no} labels (DINO) \cite{caron2021emerging}, have demonstrated remarkable success in learning rich, transferable representations from large-scale image data \cite{schmude2024prithvi, szwarcman2025prithvi, hu2025dino}. MAEs excel at learning structural patterns by reconstructing images from partially masked inputs, while DINO enables the learning of semantic features through discriminative learning \cite{caron2021emerging}. While these architectures have shown promise individually, their combined potential for satellite data analysis presents an exciting path forward, as most existing approaches focus on either local or global patterns in isolation, underutilizing the crucial interactions between different atmospheric scales. 

In this study, we present the GAIA Foundation Model, a hybrid self-supervised learning architecture that integrates MAE and DINO frameworks to learn representations that simultaneously capture local spatial structure and long-range spatial dependencies from geostationary satellite imagery. This hybrid formulation promotes the emergence of semantically meaningful latent representations, yielding enhanced generalization and superior performance on downstream meteorological tasks including precipitation retrieval, atmospheric river segmentation, and tropical cyclone instance segmentation relative to an MAE baseline.
\section{Related Work}

We organize our review of related work into three main areas: self-supervised foundation models that form the basis of our approach, deep learning methods for satellite image analysis, and specific applications in atmospheric science and remote sensing including gap filling and precipitation estimation. This structure highlights both the technical foundations and domain-specific challenges that motivate our work.

\subsection{Self-Supervised Foundation Models}
Recent advancements in self-supervised learning for computer vision enable models to learn meaningful representations without manual annotations that transfer well to domain-specific objectives \cite{vincent2010stacked}. For instance, autoencoders learn characteristic properties of the input space by compressing data into lower-dimensional latent representations and learning to reconstruct it. Early applications of convolutional autoencoders to satellite imagery demonstrated their potential for feature extraction, reconstruction, and anomaly detection \cite{romero2015unsupervised}, establishing that learned hierarchical representations can preserve geophysically meaningful structure.

MAEs extend traditional autoencoders through self-supervised pre-training on partially masked inputs \cite{he2022masked}. MAE employs an asymmetric encoder-decoder architecture based on Vision Transformers (ViT), where the encoder processes only visible patches through self-attention mechanisms, while a lightweight decoder learns to reconstruct the masked portions of the image from the unmasked embeddings. By masking a large portion of the image (typically 75\%) and minimizing reconstruction loss, MAE encourages the encoder to learn global, semantically meaningful features that transfer effectively to downstream tasks. This masking strategy promotes robustness to missing data by training the model to infer absent information- a valuable property for satellite imagery with frequent gaps.

DINO represents a complementary approach to self-supervised learning through knowledge distillation \cite{caron2021emerging}. DINO uses a student-teacher architecture where both networks share a ViT backbone but maintain distinct parameter sets—the teacher's parameters are an exponential moving average of the student's. The model learns by matching student predictions on augmented image views to teacher predictions, employing a local-to-global correspondence strategy that captures semantic relationships without explicit supervision \cite{caron2021emerging}. DINO's learned features exhibit strong properties for semantic correspondence and scene understanding, making them valuable for tasks requiring contextual interpretation \cite{caron2021emerging}.

Recent advances in foundation models have demonstrated the effectiveness of hybrid architectures that combine multiple learning paradigms \cite{moor2023foundation}. These hybrid approaches, particularly those that integrate masked auto-encoding with other self-supervised learning techniques, have shown remarkable success in learning robust and transferable representations \cite{chen2023mixed, wang2022repre}. The combination of local and global attention mechanisms, coupled with hierarchical feature learning, enables these models to capture both fine-grained details and broader contextual information. This architectural synergy has proven particularly advantageous in handling complex visual data with varying scales and temporal dependencies, making these models well-suited for atmospheric science applications. 

\subsection{Deep Learning Methods for Satellite Image Analysis}
Traditional approaches to satellite image processing have relied on physical models and statistical methods \cite{wielicki2013achieving}. However, the complexity of atmospheric phenomena and the high dimensionality of satellite data have motivated the development of deep learning-based solutions \cite{reichstein2019deep}. Recent advances in computer vision have led to significant improvements in satellite data processing, particularly in handling multi-spectral and multi-temporal data. For example, gap filling tasks have benefitted from various deep learning architectures that handle missing data in satellite observations. These include convolutional neural networks for spatial interpolation, attention-based models for capturing long-range dependencies \cite{zhang2018missing}, and physics-informed neural networks that incorporate domain knowledge \cite{wang2023toward}.



Self-supervised learning has gained traction in remote sensing as a means to leverage vast amounts of unlabeled satellite imagery \cite{jean2019tile2vec}. Various pretext tasks have been explored, including temporal prediction, spatial reconstruction, and multi-modal alignment between different satellite sensors \cite{ayush2021geography}. These methods learn representations that can transfer to downstream tasks with limited labeled data, an important consideration given the cost of expert annotation in atmospheric science. However, most existing self-supervised methods for satellite data focus on single pretext tasks and lack the flexibility to learn representations that generalize across multiple downstream applications \cite{moor2023foundation}. Additionally, computational constraints often force trade-offs between spatial resolution and temporal coverage when processing global-scale satellite data \cite{kurth2018exascale}, and integrating physical consistency constraints into self-supervised frameworks remains an open challenge \cite{reichstein2019deep}.


\subsection{Applications to Atmospheric Phenomena}


Our work focuses on four critical applications in atmospheric science: gap filling in satellite observations, instance segmentation of tropical cyclones, semantic segmentation of atmospheric rivers, and precipitation estimation. All tasks require robust feature learning and the ability to handle missing or incomplete data.

Satellite observations frequently contain missing data due to sensor limitations, cloud contamination, or transmission errors, creating gaps that can compromise downstream analyses and forecasting systems \cite{zhang2018missing}. Traditional interpolation methods, while computationally efficient, often fail to capture the complex spatiotemporal patterns and physical constraints inherent in atmospheric processes \cite{zhang2018missing}. Recent deep learning approaches have shown promise by learning spatial and temporal correlations directly from data. Convolutional neural networks (CNNs) can perform spatial interpolation, while attention-based architectures capture long-range dependencies \cite{zhang2018missing}. However, these methods typically require task-specific training and struggle to generalize across different atmospheric variables or conditions. The implicit robustness to missing data offered by masked autoencoder pre-training presents a natural fit for this problem, though its application to satellite imagery remains underexplored.

Accurate detection and tracking of tropical cyclones (TCs) is critical for disaster preparedness and climate monitoring. Traditional TC detection methods rely on manual analysis of satellite imagery or rule-based algorithms using wind speed thresholds and pressure patterns \cite{dvorak1984tropical}. The International Best Track Archive for Climate Stewardship (IBTrACS) provides a comprehensive global database of historical TC tracks and intensities \cite{knapp2010international}, which has become the standard reference dataset for training and evaluating automated TC detection systems \cite{hennon2015cyclone}. Modern deep learning approaches have significantly improved TC detection capabilities through convolutional neural networks and object detection frameworks \cite{pradhan2017tropical, wimmers2019using}. Instance segmentation methods, particularly Mask R-CNN and its variants, have been adapted for identifying individual TC instances in satellite imagery, enabling precise delineation of storm boundaries and tracking of multiple simultaneous systems. Recent work has explored the use of attention mechanisms and transformer architectures to better capture the multi-scale structure of tropical cyclones, from the compact eye region to the extensive outer spiral bands \cite{bi2022pangu}. However, these approaches often require substantial amounts of labeled training data and struggle with rare or unusual TC morphologies \cite{wang2021tropical}. Foundation models offer the potential to learn generalizable representations of TC structure from unlabeled satellite data, improving detection of developing systems and unusual storm configurations that are underrepresented in labeled datasets.

Atmospheric rivers (ARs) are elongated corridors of concentrated water vapor transport that play a crucial role in global precipitation patterns and can trigger extreme precipitation events and flooding \cite{ralph2017atmospheric, dettinger2011atmospheric}. Detecting and segmenting ARs from satellite observations is essential for understanding their climatology and improving forecasts of associated precipitation \cite{guan2015detection}. Traditional AR detection algorithms use integrated vapor transport (IVT) thresholds applied to reanalysis data or satellite-derived products \cite{rutz2014climatological, shields2018atmospheric}, but these methods often struggle with the complex, evolving morphology of ARs and may miss weaker or nascent systems \cite{ralph2020west}. Deep learning approaches for semantic segmentation, including U-Net architectures and their variants, have shown promise in delineating AR boundaries from satellite imagery by learning spatial patterns of water vapor distribution \cite{kashinath2021physics}. Recent work has incorporated multi-spectral satellite observations and temporal context to improve AR identification \cite{nguyen2023climax, kashinath2021physics}. However, the highly variable geometry of ARs—ranging from narrow filaments to broad plumes—and their interaction with complex terrain pose ongoing challenges \cite{ralph2020west}. Self-supervised foundation models that learn robust representations of atmospheric moisture patterns could enhance AR segmentation, particularly for identifying AR landfall events and tracking their evolution across ocean basins.

Estimating precipitation from satellite imagery is a fundamental task in meteorology and climate science, supporting applications from weather forecasting to water resource management \cite{hong2004precipitation}. Traditional approaches relied on infrared brightness temperature thresholds and empirical relationships \cite{joyce2004cmorph}, but modern operational systems increasingly leverage machine learning to integrate multiple satellite sensors \cite{huffman2023imerg}. The Precipitation Estimation from Remotely Sensed Information using Artificial Neural Networks (PERSIANN) family of algorithms exemplifies this evolution, progressing from simple neural networks to sophisticated deep learning architectures that combine infrared and microwave observations. Despite these advances, precipitation estimation remains challenging due to the spatial and temporal heterogeneity of precipitation patterns, varying sensor characteristics, and the need to maintain physical consistency. Foundation models that learn rich representations from unlabeled satellite data offer the potential for improved precipitation estimation, particularly in data-sparse regions or for rare precipitation events where labeled training data are limited.

While foundation models have transformed computer vision and self-supervised learning has shown promise in remote sensing, a significant gap remains in developing architectures specifically designed for atmospheric satellite data. Most existing work applies general-purpose computer vision models to satellite imagery without accounting for the unique structure of atmospheric phenomena—such as the multi-scale nature of weather systems, the physical relationships between atmospheric variables, and the systematic patterns of missing data. Furthermore, existing approaches typically focus on individual tasks in isolation, requiring separate models for gap filling, tropical cyclone detection, atmospheric river segmentation, and precipitation estimation. Our work addresses these gaps by developing a hybrid foundation model that combines the complementary strengths of masked auto-encoding and self-distillation, tailored to the characteristics of geostationary satellite observations. We demonstrate that a single pre-trained foundation model can provide robust representations that transfer effectively across all four downstream tasks, offering a unified framework for operational meteorology that reduces the need for task-specific architectures and extensive labeled datasets for each application.

\section{Methodology and Datasets}


\subsection{Data}

\begin{figure}[!htbp] 
    \centering
    \includegraphics[width=\textwidth]{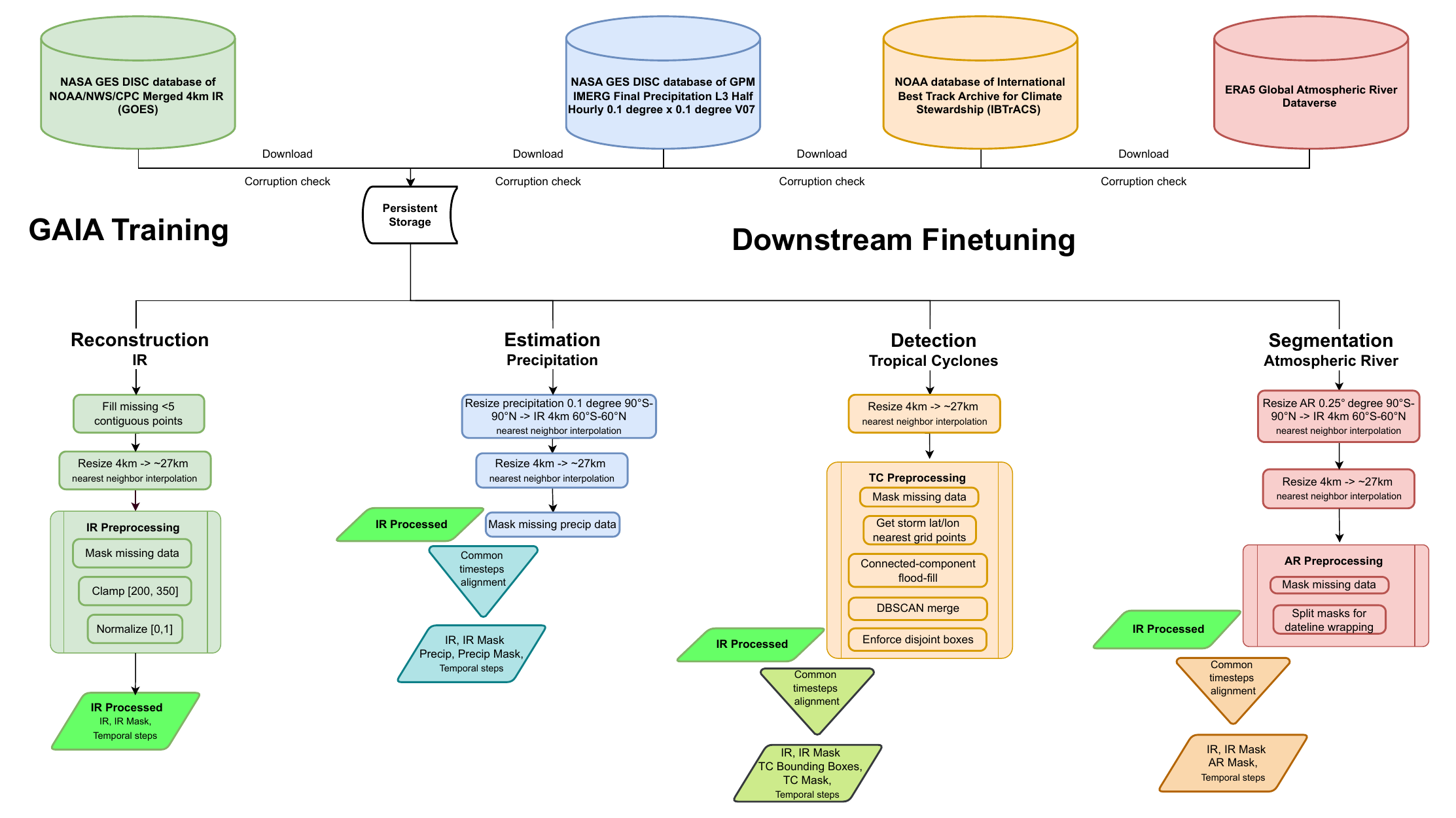}
    \caption{Data preprocessing pipelines for GAIA and downstreams, including data ingestion, cleaning, normalization, and feature extraction.}
    \label{fig:pipeline-diagram}
\end{figure}

\subsubsection{GOES Satellite Imagery}
The GOES dataset provides high-resolution, high-temporal-frequency satellite imagery critical for weather forecasting, climate monitoring, and atmospheric research. Operated by NOAA, the GOES series delivers data across multiple spectral bands, including visible, infrared, and water vapor channels. In this work, we use brightness temperature data from the cloud top temperature product, primarily derived from GOES Channel 13 (10.3 µm), part of the longwave infrared band. To achieve global coverage, we rely on the NOAA CPC Global Full Resolution Infrared Data, which incorporates GOES, METEOSAT, and Himawari data \cite{noaa_gpcp_ir, yang2016satellite}. The native dataset contains imagery at 4 km spatial resolution with snapshots of the globe every 30 minutes. Prior to training, we downscale the images to 480\(\times\)1440 pixels (approximately 27 km per pixel or 0.25 degrees) to reduce computational complexity while retaining sufficient spatial information. Even after downscaling, the imagery remains high-dimensional and structurally rich, highlighting the need for a powerful foundational model to fully interpret it.

Our preprocessing pipeline involves a two-step approach to handle missing data. First, we apply a bidirectional fill using linear interpolation within a 5-pixel radius in both latitude and longitude directions to address small voids in the data. This initial step helps maintain spatial continuity for minor gaps. Larger missing regions that cannot be filled through this local interpolation are preserved and later incorporated into our masking strategy during training.

Despite its richness, the GOES dataset frequently includes missing regions caused by sensor dropouts, calibration issues, or limited scene coverage. On average, about 7.5\% of pixels are missing in GOES imagery.\footnote{For 1000 randomly sampled images between 2005 and 2015, the average missing data amount of 7.5\% of pixels. Note that this is calculated after our data-preprocessing steps, which includes filling missing data pixels with the nearest non-missing pixel if there is a non-missing pixel within 5 pixels vertically or horizontally. So, the missing data amount in the raw data is greater than this.} An important objective of GAIA is to fill in the missing gaps in GOES data using its understanding of global weather dynamics.

\subsubsection{Downstream Task Labels}

Precipitation estimation uses ground truth data from NASA's Integrated Multi-satellitE Retrievals for GPM (IMERG) Final Precipitation product \cite{huffman2023imerg}. IMERG provides global precipitation estimates at 0.1-degree spatial resolution and 30-minute temporal resolution by combining observations from multiple satellites, including passive microwave sensors and infrared observations from geostationary satellites. We spatially aggregate the IMERG data to match our GOES resolution of 0.25 degrees using conservative remapping to preserve precipitation mass. The IMERG product provides precipitation rates in mm/hr, which serve as our regression targets for the precipitation estimation task.

To perform atmospheric river segmentation, we leverage a global AR dataset derived from applying the version 4 tARget algorithm to ECMWF Atmospheric Reanalysis v5 (ERA5) data at 6-hour intervals with a horizontal resolution of 0.25° × 0.25° \cite{guan2024regionally}. ERA5 provides comprehensive global coverage on a 1440 × 721 longitude–latitude grid. The tARget algorithm identifies ARs based on integrated water vapor transport and dynamically tracks AR features over time, incorporating enhancements to better capture zonal ARs as well as ARs in tropical and polar regions. This dataset captures multiple AR instances per timestep, and highlights each pixel with a binary (AR, no AR) label, which we use to train our segmentation model.


For tropical cyclone detection, we utilize a gridded TC region of interest (ROI) dataset that fuses the IBTrACS \cite{knapp2010international} with infrared (IR) longwave channel observations from geostationary satellites \cite{asanjan2020boundary, arellano2024tropical}. IBTrACS provides comprehensive records of tropical cyclone tracks, including center locations, intensity estimates, and temporal evolution. The gridded TC ROI product has high spatial (0.04°×0.04°) and temporal (30-minute) resolutions with global coverage, enabling precise alignment with our GOES observations. To create training labels suitable for instance segmentation, we generate bounding boxes centered around each TC location in the gridded ROI dataset. For each GOES timestamp, we identify active tropical cyclones by querying the gridded TC ROI data within a ±15-minute temporal window. For each active TC, we extract the storm center location from the high-resolution grid and convert it to pixel coordinates in our downsampled GOES grid (0.25°). Bounding box sizes vary based on the storm's maximum sustained wind speed as a proxy for spatial extent—storms with higher wind speeds receive larger bounding boxes to capture their broader cloud structure. Instance masks for each TC are created using a radial distance criterion from the storm center, where pixels within a specified radius (scaled by intensity) are labeled as belonging to that cyclone instance. This approach yields a dataset with multiple TC instances per image during active periods, particularly in peak season months.



\subsection{Model Architecture}
\label{sec:arch}

\begin{figure}[!htbp] 
    \centering
    \includegraphics[width=0.7\linewidth]{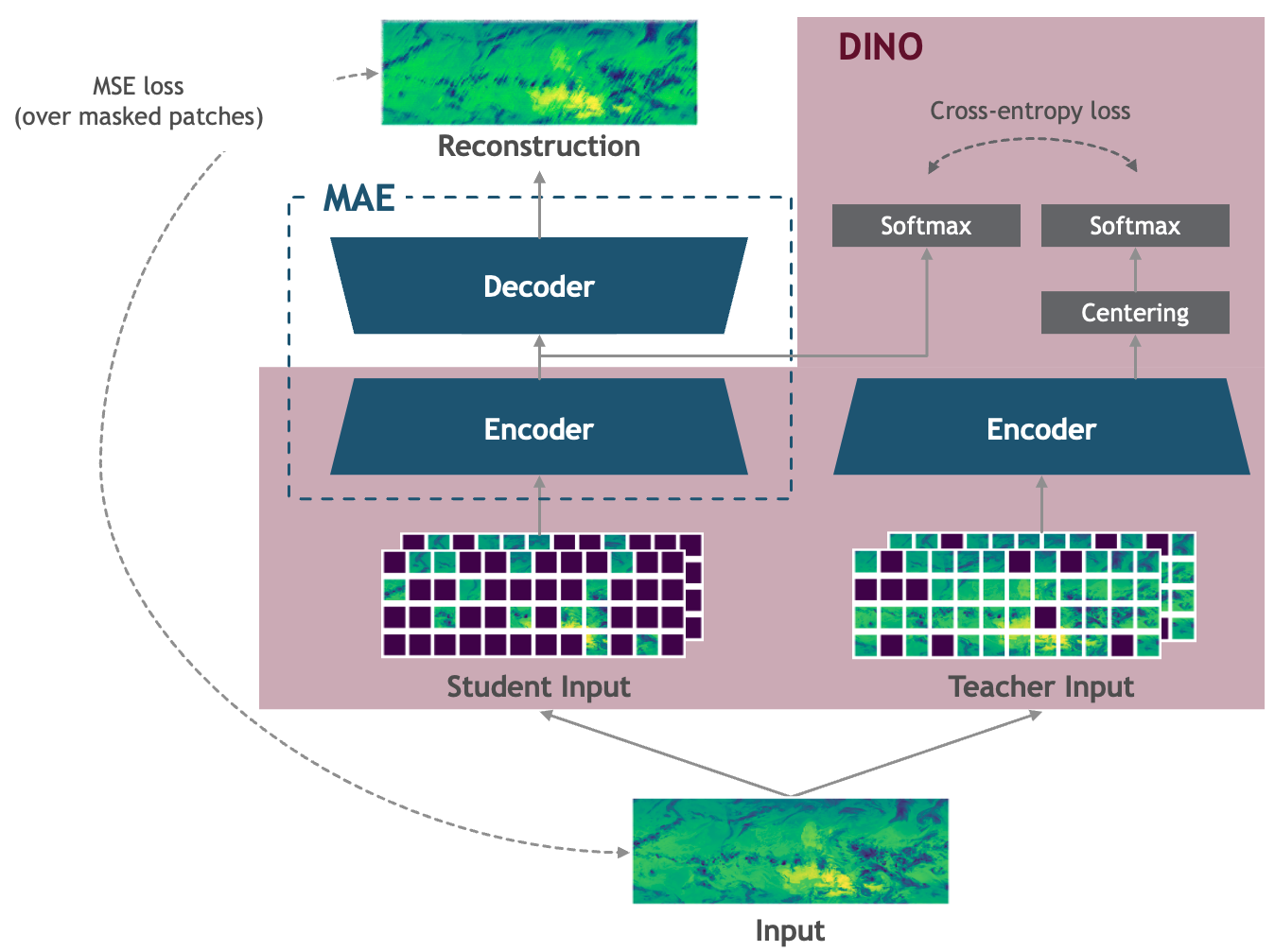}
    \caption{Architecture overview of the GAIA framework. The model combines two powerful self-supervised learning approaches: (a) MAE, which randomly masks patches of the input image and learns to reconstruct the missing regions, and (b) DINO, which uses a teacher-student architecture for knowledge distillation. The encoder processes visible patches (25\% of image for the student / MAE and 75\% of image for the teacher) through a series of transformer blocks, while the lightweight decoder reconstructs the full image from visible tokens. This architecture enables efficient self-supervised pretraining on large-scale datasets without requiring manual labels, leading to robust visual representations that can be fine-tuned for downstream tasks.}
    \label{fig:mae-dino-diagram}
\end{figure}

Our hybrid architecture, GAIA (illustrated in Figure \ref{fig:mae-dino-diagram}), integrates the MAE and DINO frameworks through a shared encoder, enabling simultaneous learning of local and global features. The architecture consists of three main components:

\begin{enumerate}
    \item A shared ViT-based encoder that processes patches for both MAE and DINO branches
    \item A lightweight decoder for reconstruction in the MAE branch
    \item A momentum-updated teacher network for self-distillation in the DINO branch
\end{enumerate}

The encoder follows the ViT architecture with an embedding dimension of 912 (matching our 30\(\times\)30\(\times\)1 patch size), 24 layers, and 16 attention heads. The decoder is more lightweight, with an embedding dimension of 512, 8 layers, and 16 attention heads. This asymmetric design allows for efficient computation and ensures that  reconstruction quality is driven by representative embeddings from the encoder.

\subsection{Training Strategy}
We jointly pre-train GAIA using a convex combination of the self-distillation (DINO) and reconstruction (MAE) objectives. At epoch $\epsilon$ the loss is computed as follows:

$$\L_{\epsilon} = \lambda(\epsilon) \times \L_{DINO} + (1 - \lambda(\epsilon)) \times \L_{MAE}$$

Where $\lambda$ follows the below piecewise linear schedule:

\begin{equation}
\lambda(\epsilon) =
\begin{cases}
1, & 0 \le \epsilon < E_w, \\[4pt]
1 - \dfrac{\epsilon - E_w}{E_p}\,\bigl(1 - \lambda_\star\bigr), & E_w \le \epsilon < E_w + E_p, \\[8pt]
\lambda_\star, & \epsilon \ge E_w + E_p ,
\end{cases}
\label{eq:lambda-schedule}
\end{equation}

\noindent where $E_w$ is the DINO warm-up length, $E_p$ the transition length,
and $\lambda_\star$ the final DINO loss weight. For our pre-training, we leverage a 5-epoch DINO warm-up followed by a 20-epoch transition phase, linearly introducing the MAE loss at each epoch. In this framework, early training focuses on invariant feature discovery (DINO) without the reconstruction signal. We then linearly introduce the MAE loss to encourage spatially grounded features while retaining biases learned by DINO. The final phase keeps a fixed mixture of the DINO and MAE objectives to stabilize optimization until convergence. We found in early training attempts that this framework mitigates the risk of DINO collapse in the case when the model optimizes too heavily on the reconstruction objective and is unable to find a non-trivial optimum for the DINO objective.

We induce global–local alignment within the DINO component by employing an asymmetric masking scheme between the teacher and student networks. We leverage masking as the source of augmentation for DINO (rather than the multi-crop used in \cite{caron2021emerging}) to enable parameter sharing between the MAE and DINO components. This choice is further justified by noting that we expect identically-sized inputs in satellite data; hence, invariance to crops and rotations are less salient to our data distribution. The DINO teacher produces output for two views with 25\% masking while the student produces two views using 75\% masking. This diversity compels the student to align local cues with the teacher’s more global view, preserving DINO’s benefits in a masked setting. We leverage grid-search to identify suitable hyperparameters to mitigate the risk of DINO collapse; we update the teacher network in the DINO branch with a momentum coefficient of 0.996 and update the DINO center with a multiplicative momentum of 0.9.


Under this framework, we train GAIA over 15 years of GOES data (2001-2015) at 0.25 degree resolution, optimizing memory usage via \texttt{Bfloat16} precision.

\subsection{Hardware and Computational Requirements}

The GAIA model was trained and evaluated on the National Research Platform (NRP), a distributed Kubernetes cluster accessible across U.S. universities.

The computational infrastructure utilized NVIDIA A10 GPUs located at UNL (University of Nebraska-Lincoln), with the training process configured for distributed training across multiple nodes. Key training parameters were managed through a Kubernetes job specification, allowing for flexible allocation of resources.

GAIA was trained on 5 nodes, each with 8 NVIDIA A10 GPUs (40 in total), using DeepSpeed Stage 2 for distributed training in bf16 mixed precision. The global batch size was 120 (with a micro-batch of 3 per GPU and 1 gradient accumulation step). We used the AdamW optimizer with an initial learning rate of 5e-4 and applied a cosine annealing scheduler. The GAIA base model was trained for over 2 weeks to convergence.

\subsection{Downstream Tasks}


We evaluate the generalizability of GAIA's embeddings across three representative downstream tasks: precipitation estimation, atmospheric river segmentation, and tropical cyclone detection. These tasks both represent diverse objectives -- dense regression, segmentation, and event detection -- while also being fundamental challenges in atmospheric modeling, making them suitable tests for GAIA.




\subsubsection{Precipitation Estimation}





We cast the precipitation task as a regression from infrared imagery to a per-pixel precipitation rate (mm/hr), which presents unique challenges due to the nonlinear relationship between infrared brightness temperatures and precipitation rates. For this, we retain the base model encoder and extend the attention-based decoder used for base model training in the following ways: 

\begin{itemize}
    \item[i] Adding a ReLU activation in the output layer, ensuring physically meaningful (non-negative) precipitation estimates
    \item[ii] Fine-tuning to minimize mean squared error over the full (precipitation) image
    \item[iii] Leveraging a small (10\%) mask ratio to ensure robust performance in the presence of missing data
\end{itemize}



\subsubsection{Atmospheric River Segmentation}

The atmospheric river segmentation task presents two unique challenges in comparison to the precipitation task:

\begin{itemize}
    \item[i] ARs are coherent structures of varying scales, detecting them requires consistent classification between neighboring pixels
    \item[ii] They are represented as discrete masks, rather than continuous pixel-level values, motivating important architectural adaptations relative to the precipitation decoder
\end{itemize}

In light of the above challenges, we formulate the task as a pixel-level binary classification (AR vs. background) problem. For this task, we mirror the precipitation decoder but replace the ReLU activation layer with a single-channel classification head that produces logits per pixel. The model is trained using a binary cross-entropy (BCE) with logits objective, and leverages a sigmoid activation at inference time to convert logits to probabilities. The transformer decoder’s long-range attention, coupled with the single-channel head and logits-based training, encourages clear separation between AR and background pixels while remaining computationally light. We train the AR model on three years of data (2013-2015) and evaluate on 2020 data to evaluate the model's ability to generalize to out-of-sample data.


\subsubsection{Tropical Cyclone Detection}

We formulate tropical cyclone detection with GOES data as an instance-level segmentation problem; each storm represents a small, coherent structure to be identified via a bounding box, and specifically delineated with binary masks. For this task, we adopt the Mask-RCNN \cite{he2017mask}, which provides several advantages within our framework:

\begin{itemize}
    \item The joint outputs—bounding boxes and pixel-wise masks—enable coarse-to-fine spatial understanding: boxes support robust tracking and scene context, while masks unlock per-pixel analyses (shape, temperature) that characterize specific cyclone structure and potential evolution
    \item The flexibility of the architecture allows us to plug in our GAIA encoder as the model's backbone and fine-tune using the detection heads
    \item The two-stage detection framework, with a region proposal network (RPN) and ROI head, allows us to both maintain high recall despite the large class imbalance typical of global TC data while still preserving a strong sense of spatial locality
    \item The model naturally integrates with feature pyramids for multi-scale detection, which is important due to the variability in TC sizes across the globe
\end{itemize}

We initialize the Mask-RCNN backbone using GAIA, leaving the model unfrozen to allow for full adaptation to the detection task. To provide scale-specific features, we reshape the patch tokens into a 2-dimensional grid at the encoder's patch resolution, apply a lateral, $1{\times}1$ projection to $256$ channels, and follow with a $3{\times}3$ convolution to introduce additional locality; this forms our base feature map $P_0$. We then upsample $P_0$ into a lightweight multi-scale feature pyramid via transposed convolutions, producing $P_i$ at finer resolutions, where the resolution of $P_i$ = $2 \times P_{i-1}$. The final pyramid, built from a single source, consists of 4-levels, and plays the role of an FPN, easing the burden of multi-scale detection while remaining largely reliant on GAIA's learned representation.

The methodology of the detector component leverages the standard Mask-RCNN implementation. The RPN operates on each $P_i$ to propose candidate regions, where proposals are assigned to pyramid levels by size. The ROI head then performs finer classification and bounding box regression on these proposals to better align them to true cyclones, and the mask branch predicts a binary mask per instance, obtaining footprints for predicted cyclones. The training objective follows the traditional Mask-RCNN objective:

$$\L = \L_{rpn-cls} + \L_{rpn-reg} + \L_{box-reg} + \L_{cls} + \L_{mask}$$

We additionally tune the anchor generator and box ROI poolers to our task, and reduce the Intersection over Union (IoU) threshold for positive examples to favor a higher concentration of positive examples for training. 

\section{Experiments and Results}
\subsubsection{Principal Components Analysis}

To investigate how our dual objectives influence the encoder, we perform Principal Component Analysis (PCA) on per-timestep embeddings and analyze the top 3 principal components of GAIA against baseline MAE- and DINO-only models. To isolate within-image variation, the embedded images are individually de-meaned prior to decomposition. Finally, patch-level embeddings are upsampled to pixel resolution using AnyUp \cite{wimmer2025anyup} for visualization. 

\begin{figure}[htbp]
    \centering
    \includegraphics[width=0.95\textwidth]{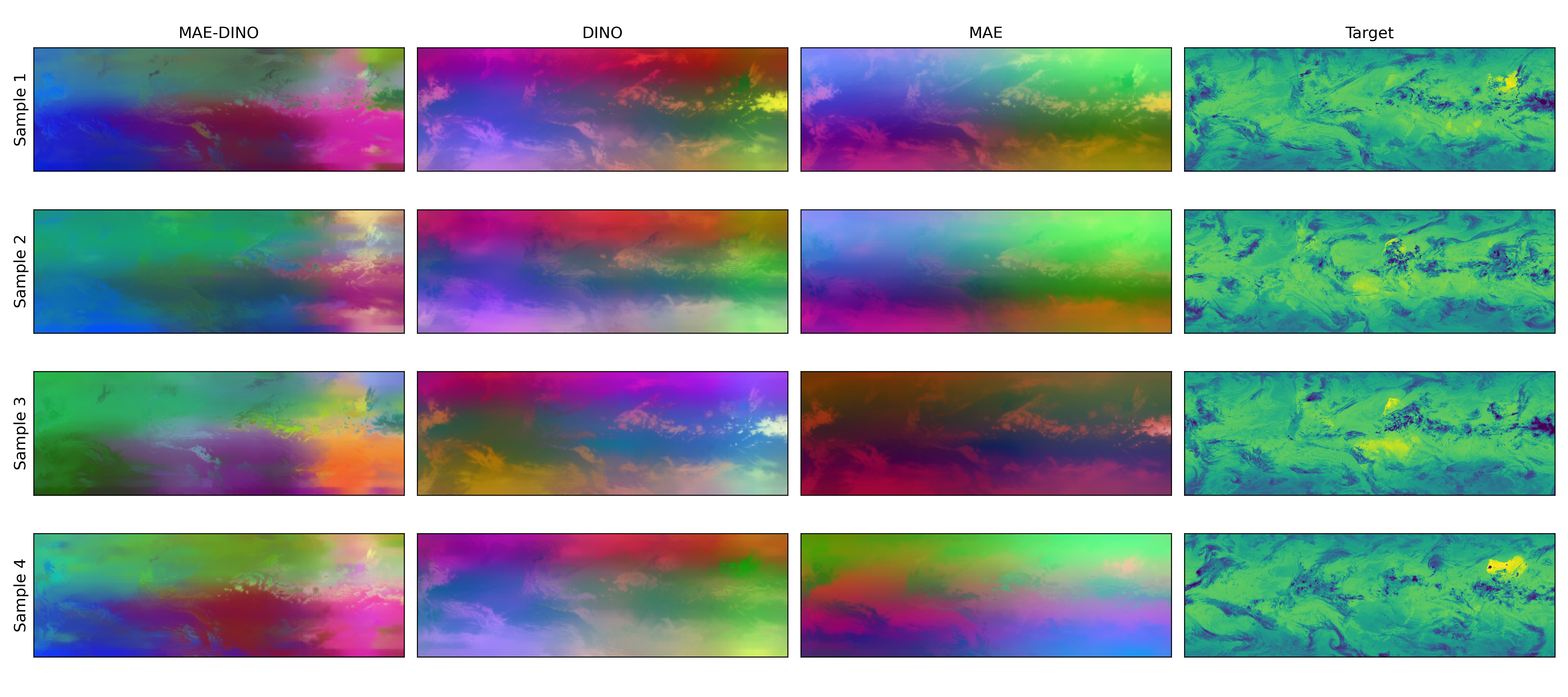}
    \caption{PCA projections of learned patch embeddings across four sample timesteps. PCA performed on 912-dimensional embedding vectors produced by the encoder. The RGB channels correspond to principal components 1, 2, and 3, respectively. From left to right: GAIA (MAE-DINO), DINO-only, MAE-only, and target IR input data. Patch-level embeddings are upsampled to pixel resolution using AnyUp \cite{wimmer2025anyup} for visualization. GAIA (first column) shows distinct pattern separation with spatially coherent features aligned to meaningful atmospheric zones (e.g., equatorial cloud belts, longitudinal structures). In contrast, DINO (second column) and MAE (third column) produce embeddings that closely resemble smoothed versions of the input (fourth column) with positional structure, indicating concentrated, reconstruction-focused representations rather than disentangled semantic features.}
    \label{fig:pca_proj_rand}
\end{figure}

We display our results in Figure \ref{fig:pca_proj_rand} where the principal components are assigned to the red, green, and blue channels respectively. In addition, we note the explained variance for each of the top 3 principal components in Table \ref{tab:explained_variance_pca_rand}. We observe that MAE- and DINO-only embeddings exhibit highly concentrated variance: the first three PCs explain over 88\% of total variance, with PC1 alone capturing around 80\%. Visually, these embeddings appear as smoothed versions of the input, with intensity patterns over cloud regions and broad color gradients consistent with the sinusoidal positional embeddings upstream of the encoder. This concentration suggests the single-objective models learn compressed representations that primarily encode low-frequency spatial information and positional structure—a potentially trivial solution optimized for pixel-level reconstruction (MAE) or potentially brittle satellite image fingerprinting (DINO) rather than semantic understanding.

In contrast, GAIA demonstrates a more distributed variance profile (72.5\% cumulative across PC1-3, with 42\% in PC1, 19\% in PC2, and 11\% in PC3), which we interpret as evidence of richer semantic learning. The lower but more spread-out variance indicates that the representation is not dominated by any single pattern but instead disentangles different aspects of atmospheric structure across multiple principal components. Critically, the visual patterns in GAIA embeddings reveal distinct, spatially coherent features that align with meaningful atmospheric zones—the green-yellow bands correspond to equatorial cloud belts, while vertical color bands indicate longitudinal structures. Different principal components capture distinct geographical and meteorological patterns, suggesting that similar atmospheric features are grouped together in the embedding space rather than simply replicating the input signal. This distributed variance structure indicates that the combined distillation and reconstruction objectives encourage the encoder to learn a semantically organized and disentangled representation, where information is spread across multiple informative dimensions rather than collapsed into a low-dimensional, reconstruction-focused encoding. Such representations typically prove more transferable to downstream tasks, as different tasks can leverage different subspaces of the learned embedding.

This finding aligns with observations in self-supervised vision learning, where combining complementary pretraining objectives has proven beneficial. For instance, iBOT \cite{zhou2022ibot} combines masked image modeling with self-distillation---analogous to our MAE and DINO objectives---and demonstrates that this combination yields stronger visual representations than either objective alone. Our downstream experiments (Section \ref{sec:downstreams}) reach a similar conclusion: combining the MAE and DINO objectives produces flexible embeddings better suited for diverse atmospheric modeling tasks, with GAIA consistently outperforming single-objective baselines across precipitation estimation, atmospheric river segmentation, and tropical cyclone detection.

\begin{table}[h!]
\centering
\begin{tabular}{l|ccc|c}
\hline
\textbf{Model} & \textbf{PC1 (\%)} & \textbf{PC2 (\%)} & \textbf{PC3 (\%)} & \textbf{Cumulative (\%)} \\
\hline
MAE       & 82.7 & 4.6 & 2.9 & 90.2 \\
DINO      & 79.7 & 4.9 & 3.5 & 88.1 \\
GAIA  & 42.4 & 19.2 & 10.9 & 72.5 \\
\hline
\end{tabular}
\captionsetup{skip=8pt}  
\caption{Explained variance ratio (as percentages) of PCA decomposition for the MAE-only and GAIA models, rounded to 1 decimal place.}
\label{tab:explained_variance_pca_rand}
\end{table}

\subsubsection{Temporal Embedding Analysis}

To assess whether GAIA learns meaningful atmospheric dynamics rather than trivial diurnal patterns, we analyze the temporal structure of learned embeddings. A model that merely memorizes solar illumination cycles would produce embeddings strongly correlated with time-of-day, showing high similarity among all observations at the same hour regardless of actual weather conditions. In contrast, a model that learns genuine atmospheric processes should exhibit embeddings that reflect weather variability rather than being dominated by the 24-hour solar cycle.

\begin{figure}[htbp]
    \centering
    \includegraphics[width=0.7\textwidth]{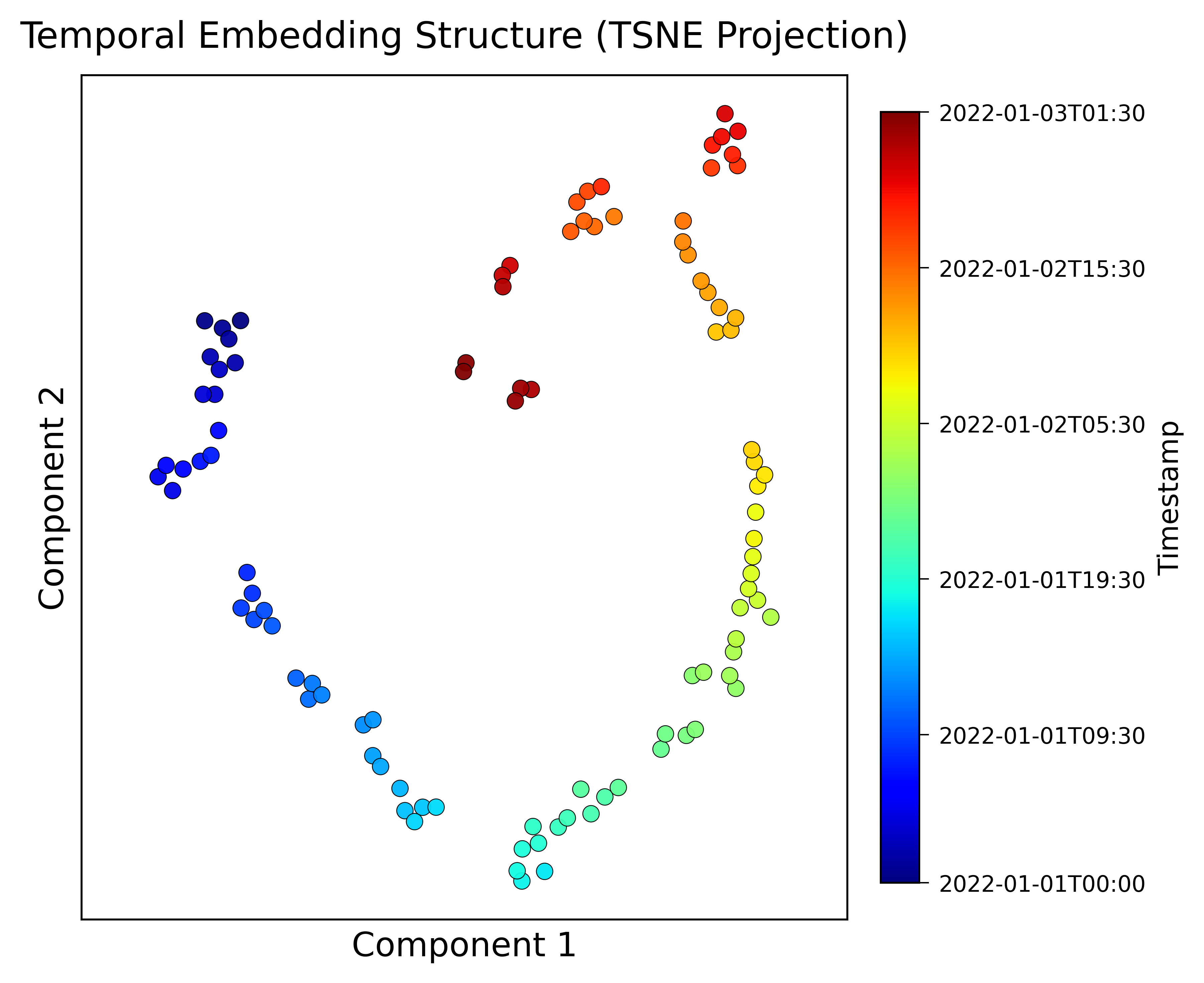}
    \caption{t-SNE projection of GAIA embeddings over 3 days (January 1-3, 2022). Each point represents a single timestep, with color indicating chronological progression. Embeddings form a smooth temporal trajectory where nearby points correspond to nearby times, demonstrating that the model learns temporally coherent representations. The continuous structure without discrete clusters indicates that embeddings reflect atmospheric evolution rather than categorical time-of-day encodings.}
    \label{fig:tsne_temporal}
\end{figure}

Figure \ref{fig:tsne_temporal} visualizes the temporal structure of embeddings using t-SNE projection over a 3-day period (January 1-3, 2022). The embeddings form a smooth, continuous trajectory through the latent space, with color indicating chronological progression from blue (early times) to red (later times). Critically, temporally adjacent observations cluster together in embedding space—images taken 30 minutes apart have similar representations, while observations separated by hours or days are more distant. This organization demonstrates that GAIA captures smooth temporal evolution of atmospheric states, where the embedding geometry reflects actual temporal proximity rather than arbitrary time-of-day labels. The continuous trajectory without abrupt jumps suggests the model learns physically consistent transitions between atmospheric configurations.

This analysis demonstrates that GAIA's learned representations capture physically meaningful atmospheric evolution. The smooth temporal trajectories in embedding space reflect genuine transitions between weather states, while the absence of strong diurnal clustering confirms that the model prioritizes cloud dynamics and atmospheric structure over simple solar illumination patterns. This property is essential for downstream tasks like cyclone tracking and precipitation estimation, which require representations sensitive to weather variability rather than time-of-day.

\subsubsection{Gap-filling Performance}
We evaluate GAIA's out-of-the-box ability to reconstruct missing regions in global satellite imagery, a critical task for handling sensor malfunctions or data transmission issues. On average, 7.5\% of the GOES dataset is affected by missing data, compromising the effectiveness of models such as weather prediction and cyclone detection models which utilize this and similar data.

\subsubsection{Quantitative Analysis}
We quantitatively evaluate GAIA's gap filling ability by synthetically masking around 3,000 samples taken from the full year of 2020 and computing the Root Mean Square Error (RMSE) between the gap-filled images from the model and the ground truth before synthetic masking. Figures \ref{fig:artificial_gap_filling} and \ref{fig:missing_gap_filling} show the regions that were synthetically masked through purple outlines on the reconstruction images. Quantitative results are shown in Table \ref{tab:gapfill_metrics}. The model's high structural similarity index measure (SSIM) and low RMSE indicates strong gapfill performance across all six synthetic masks. 

\begin{table}[h!]
\centering
\caption{Gap-filling performance metrics for 2020. We compute SSIM over the entire image and RMSE over the synthetically masked portions of the data. Best values presented in bold.}
\sisetup{
    round-mode=places,
    round-precision=4,
    table-number-alignment = center
}
\begin{tabular}{ll
                S[table-format=1.4]
                S[table-format=1.4]
                S[table-format=1.4]
                S[table-format=1.4]}
\toprule
\textbf{Experiment} & \textbf{Mask / Ratio} & \multicolumn{2}{c}{\textbf{GAIA}} & \multicolumn{2}{c}{\textbf{MAE Only}} \\
\cmidrule(lr){3-4} \cmidrule(lr){5-6}
 &  & \textbf{SSIM} & \textbf{RMSE} & \textbf{SSIM} & \textbf{RMSE} \\
\midrule
\multirow{4}{*}{Missing Data}
 & A & \textbf{0.5243} & \textbf{0.0865} & 0.4442 & 0.1104 \\
 & B & \textbf{0.5215} & \textbf{0.0960} & 0.4428 & 0.1208 \\
 & C & \textbf{0.5247} & \textbf{0.0912} & 0.4512 & 0.1091 \\
 & D & \textbf{0.5250} & \textbf{0.0959} & 0.4492 & 0.1149 \\
\midrule
\multirow{2}{*}{Artificial Masking}
 & Vertical Bars& \textbf{0.5280} & \textbf{0.0994} & 0.4690 & 0.1227 \\
 & Horizontal Bars& \textbf{0.5277} & \textbf{0.0981} & 0.4699 & 0.1228 \\
\midrule
\multirow{5}{*}{Mask Ratio}
 & 30\% & \textbf{0.5212} & \textbf{0.0906} & 0.4430 & 0.1205 \\
 & 50\% & \textbf{0.5226} & \textbf{0.0912} & 0.4551 & 0.1224 \\
 & 70\% & \textbf{0.5260} & \textbf{0.0948} & 0.4748 & 0.1231 \\
 & 90\% & \textbf{0.5321} & \textbf{0.1032} & 0.5023 & 0.1223 \\
 & 95\% & \textbf{0.5331} & \textbf{0.1078} & 0.5101 & 0.1230 \\
\bottomrule
\end{tabular}
\vspace{0.5em}
\label{tab:gapfill_metrics}
\end{table}

\begin{figure}[!htb]
    \centering
    \includegraphics[width=0.7\textwidth]{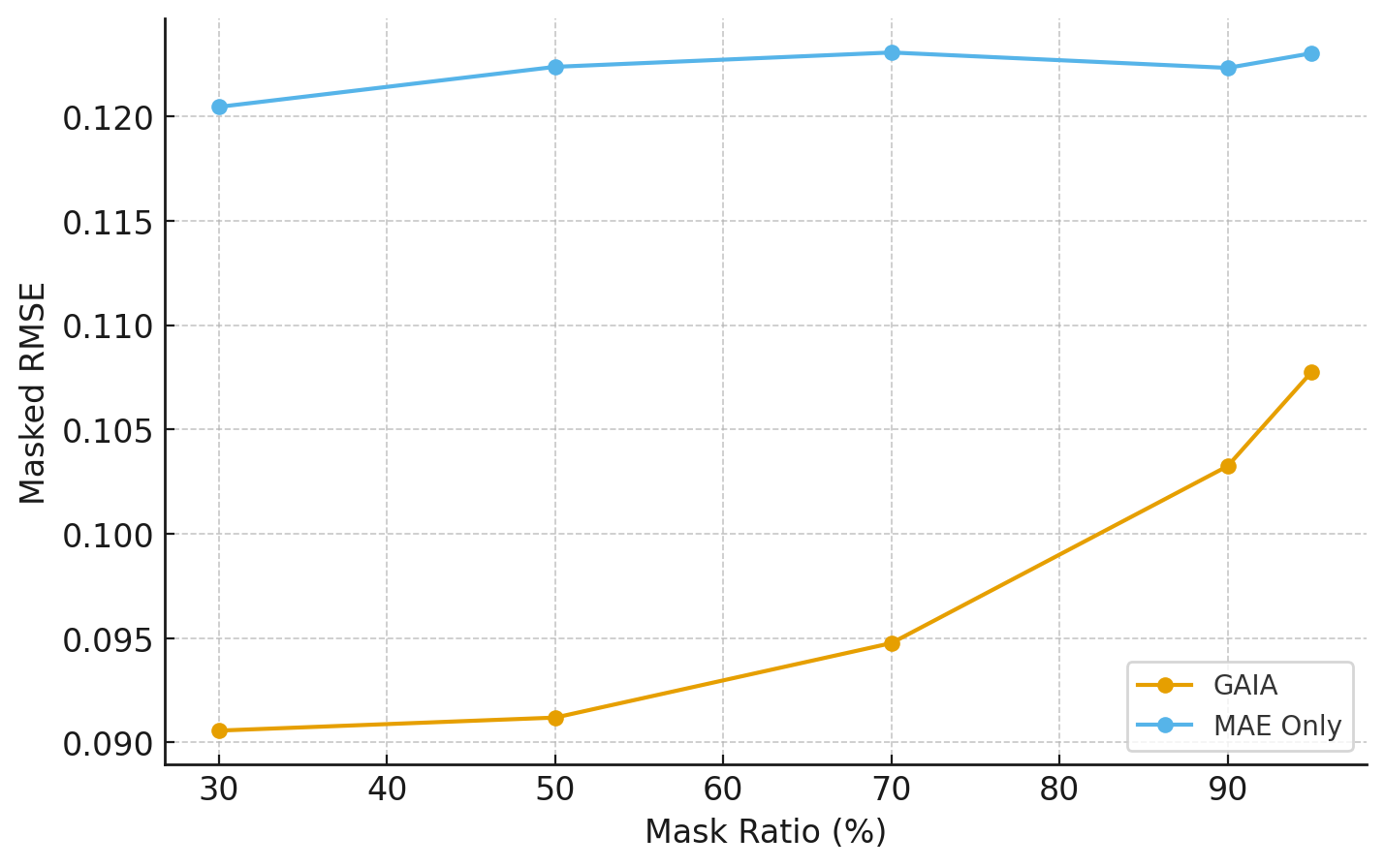}
    \caption{Quantitative evaluation of gap filling quality across different mask ratios. RMSE over masked patches is lower across the board for GAIA.}
    \label{fig:mask_ratio_metrics}
\end{figure}

We also evaluate our model by running inference with varying artificial mask ratios to test its robustness. Figure \ref{fig:mask_ratio_metrics} shows how reconstruction quality varies with mask ratio. The figure shows that as the mask ratio increases (less context available to the encoder), RMSE for the artificially masked patches increases, indicating degraded reconstruction quality. This behavior is intuitive - with more patches hidden, the model has less contextual information to guide its predictions. GAIA's encoder can extract meaningful representations from limited visible patches, a capability enabled by the model's pre-training on high mask ratios (75\%) combined with the DINO objective's emphasis on learning robust, mask-invariant features. The GAIA model experiences roughly quadratic degradation as mask ratio increases, while the MAE-only model performance degradation is roughly constant and exhibits low degradation as mask ratio increases. This suggests the MAE-only model does not utilize all of the context available to it to make its predictions. 

\subsubsection{Qualitative Analysis}
\begin{figure}[!htb]
    \centering
    \begin{subfigure}{\textwidth}
        \centering
        \includegraphics[width=\textwidth]{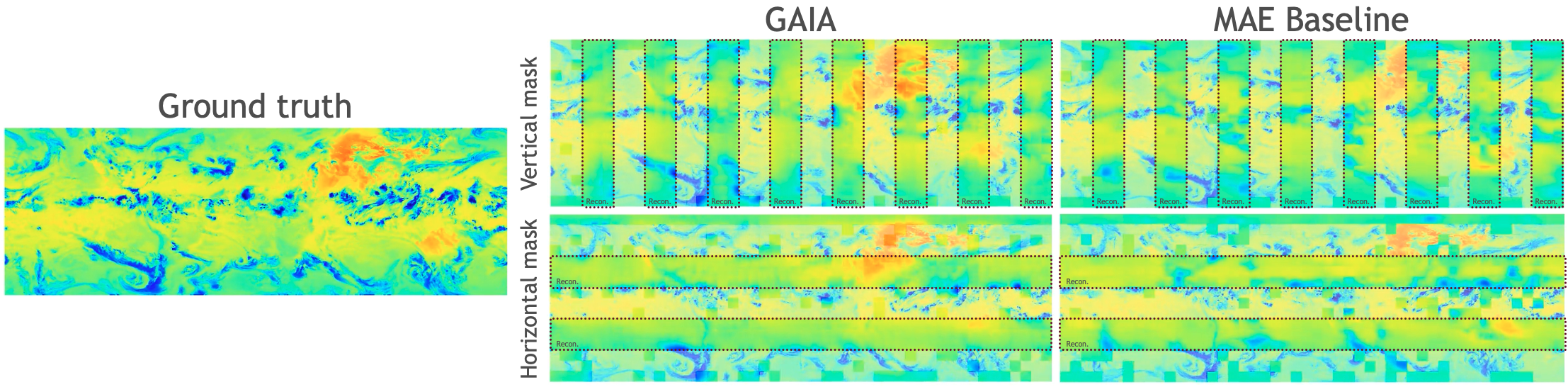}
    \end{subfigure}

    \caption{Comparison of ground truth (left), GAIA reconstruction (middle), and MAE-only reconstruction (right) for vertically and horizontally striped masking on a single timestep (June 5th, 2023 at 4:30AM). The lighter patches are the unmasked regions (with ground truth overlaid), while the rest of the image is gap-filled by the model (with purple dashes outlining where the synthetic mask was applied). The GAIA model successfully reconstructs missing regions while preserving both spatial patterns and temperature intensity distributions from the unmasked regions.}
    \label{fig:artificial_gap_filling}
\end{figure}

\begin{figure}[!htb]
    \centering
    \begin{subfigure}{\textwidth}
        \centering
        \includegraphics[width=\textwidth]{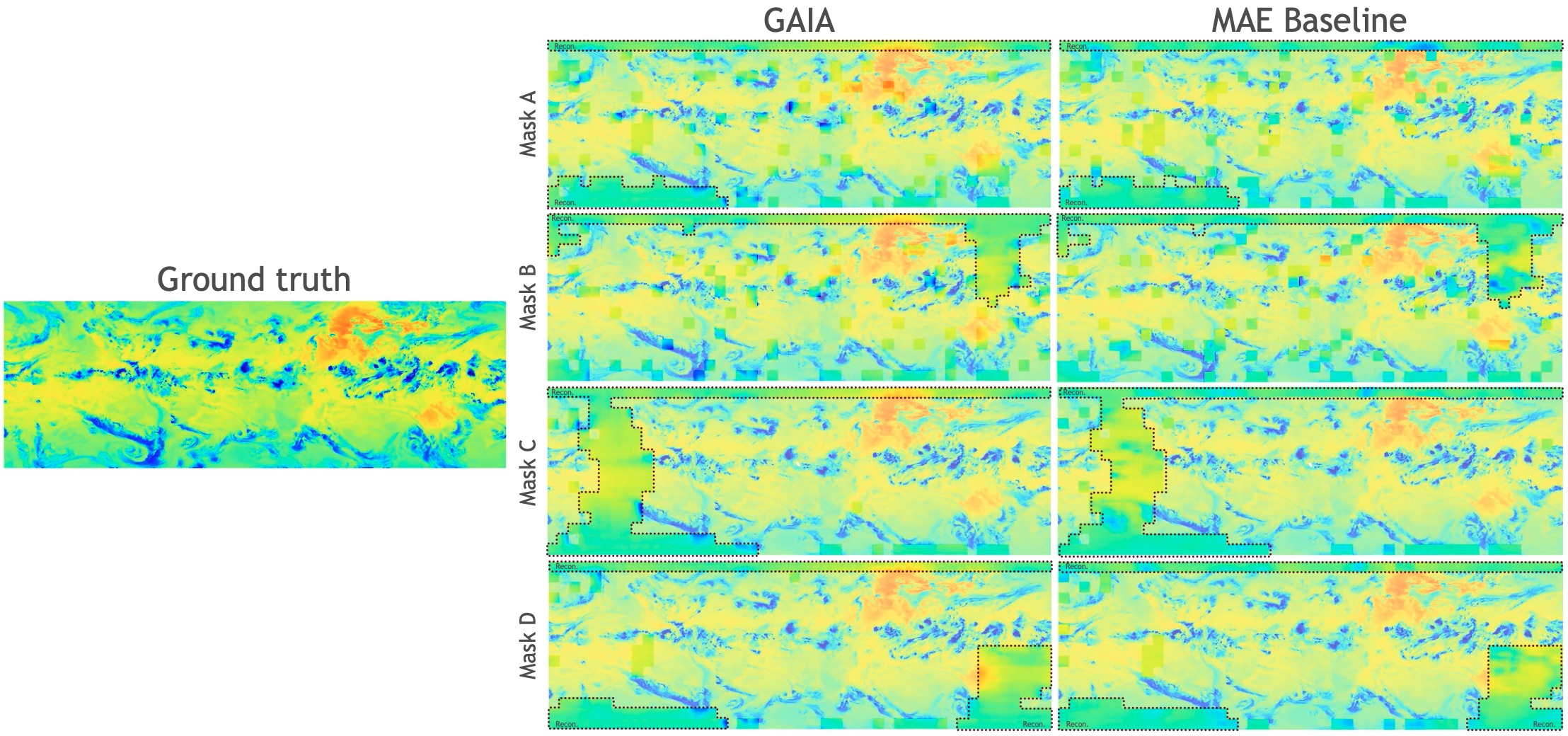}
    \end{subfigure}

    \caption{Comparison of ground truth (left), GAIA reconstruction (middle), and MAE-only reconstruction (right), for real examples of missing data pulled from other samples, on a single ground-truth timestep (June 5th, 2023 at 4:30AM). The lighter patches are the unmasked regions (with ground truth overlaid), while the rest of the image is gap-filled by the model (with purple dashes outlining where the synthetic mask was applied). The GAIA model successfully reconstructs missing regions while preserving both spatial patterns and temperature intensity distributions from the unmasked regions.}
    \label{fig:missing_gap_filling}
\end{figure}

Figures \ref{fig:artificial_gap_filling} and \ref{fig:missing_gap_filling} demonstrate the model's gap filling and reconstruction capabilities across two synthetic masks and four missing data masks extracted from the GOES dataset. Each column displays the same timestep, with the first column containing the ground truth, the second column containing the ground truth gap-filled with GAIA's predictions, and the third column containing the ground truth gap-filled with the MAE-only model's predictions.

The results showcase two key strengths of our approach:
\begin{itemize}
    \item \textbf{Large Gap Reconstruction}: The model successfully reconstructs substantial missing regions while preserving the temperature gradients and atmospheric patterns consistent with surrounding areas (e.g., GAIA results in Figure \ref{fig:artificial_gap_filling}).
    
    \item \textbf{Pattern Continuity}: The reconstructions maintain smooth transitions between filled regions and original data, avoiding artificial boundaries or discontinuities (for instance, compare the gap-filled bottom-left corners of both models' predictions in Figure \ref{fig:missing_gap_filling}).
\end{itemize}

However, there are still some key areas for improvement. In particular, fine-scale features such as cloud formations and temperature variations are not always accurately reproduced (with the exception of high-temperature features such as the orange patch gap-filled correctly in Figure \ref{fig:artificial_gap_filling}). We hypothesize this is a result of our relatively large patch size (252 sq. km per patch), combined with MSE-trained models' tendencies to ``predict the average.'' We expect training on higher resolution images and with smaller patch sizes in future iterations of the model will ameliorate these concerns.

These results demonstrate GAIA's effectiveness in handling real-world scenarios where satellite data may be partially missing or corrupted, providing a reliable solution for maintaining continuous global climate monitoring.

\subsection{Downstream Task Evaluation}\label{sec:downstreams}

\subsubsection{Precipitation Estimation Evaluation}

We trained our GAIA precipitation estimation model on a full year of data from 2005 and evaluated its performance
across a representative sample from 2020 (two weeks from each season). As a point of comparison, we trained an
identical model leveraging a baseline MAE encoder; the results of both are reported in the analysis below. We note that for precipitation estimation, the model trained from GAIA does not perform conclusively better than the one trained from only MAE. We hypothesize that this stems from the relative simplicity and input-aligned nature of the task-- precipitation levels are highly correlated with cloud-top temperatures; as such, we would expect a highly input-aligned embedding such as the one generated by MAE to perform as well as any other reasonable choice of embedding.

\subsubsection{Quantitative Analysis}
\label{sec:precip_quant}
To quantify the model's accuracy and reliability, we computed five key evaluation metrics: RMSE, patch- and pixel-level accuracy, and patch- and pixel-level false alarm ratio (FAR) as shown in Table \ref{tab:precip_metrics}. To compute patch-level accuracy and FAR metrics, we compute the mean predicted precipitation over each of the $30 \times 30$ pixel patches created by our ViT / MAE encoder, motivated by the fact that the models' embeddings are per-patch rather than per-pixel.

Both metrics of FAR and accuracy treat our precipitation task as binary classification of precipitation/no-precipitation; we threshold our predictions at 0.5 mm / hour for both the patch- and pixel-level metrics. The definitions of these metrics are shown in Equation \ref{eq:precip_metrics}.

\begin{equation}
\label{eq:precip_metrics}
\mathrm{Accuracy} = \frac{\mathrm{\#\ True\ Positive} + \mathrm{\#\ True\ Negative}}{\mathrm{\# Samples}}, \qquad
\mathrm{FAR} = 1 -  \frac{\mathrm{\#\ True\ Positive}}{\mathrm{\# Predicted\ Positive}}.
\end{equation}

\begin{table}[!htb]
\centering
\sisetup{
  parse-numbers       = true,
  round-mode          = figures,
  round-precision     = 4,
  detect-weight       = true,
  detect-inline-weight= math,
  table-number-alignment = center
}
\begin{tabular}{l S[table-format=1.8] S[table-format=1.8]}
\toprule
\textbf{Metric} & \textbf{GAIA} & \textbf{MAE Only} \\
\midrule
RMSE                 & {\bfseries \num{0.688428998}} & \num{0.698196054} \\
Patch Accuracy (\%)  & {\bfseries \num{95.5}}        & \num{95.4}        \\
FAR Patch (\%)       & {\bfseries \num{26.8}}        & \num{28.6}        \\
Pixel Accuracy (\%)  & \num{93.6}                    & {\bfseries \num{93.7}} \\
FAR Pixel (\%)       & \num{61.3}                    & {\bfseries \num{61.2}} \\
\bottomrule
\end{tabular}
\vspace{0.5em}
\caption{Precipitation estimation performance metrics for 2020. The model shows strong performance across all metrics, particularly in false alarm ratio and structural similarity, indicating reliable precipitation detection and pattern reproduction.}
\label{tab:precip_metrics}
\end{table}

Together, these evaluation metric results suggest the model generalizes well from a limited training window to out-of-sample evaluation, maintaining good accuracy, structural realism, and low false positives in its precipitation predictions. 

\subsubsection{Qualitative Analysis}

\begin{figure}[ht]
  \centering
  \includegraphics[width=\textwidth]{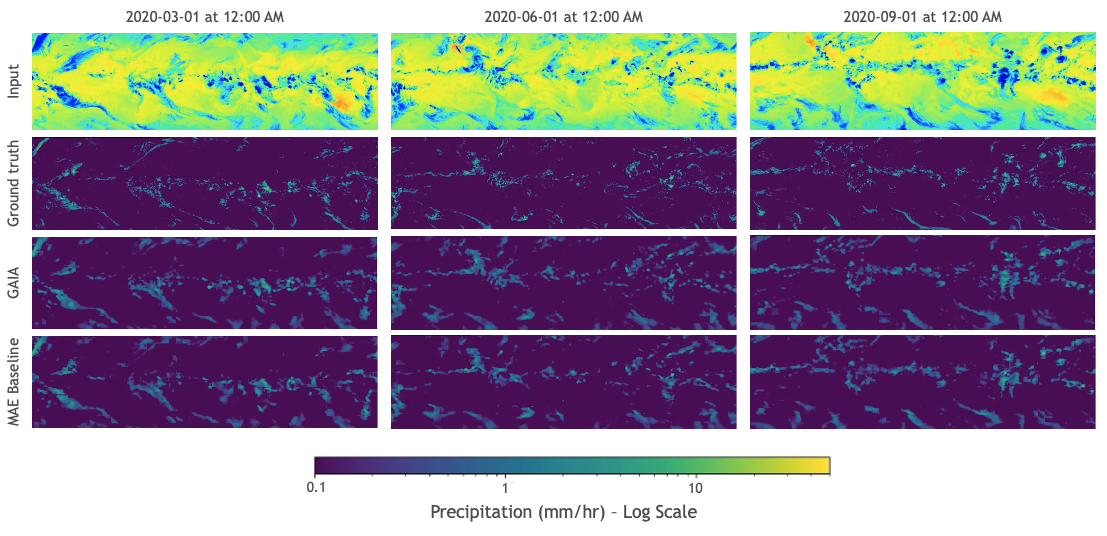}
  \caption{
Comparison of IR input (top row), ground truth (second row), and model-estimated precipitation (third and fourth rows for GAIA and MAE, respectively) for three timestamps in 2020. Each row corresponds to midnight on a specific date (left to right: March 1, 2020; June 1, 2020; September 1, 2020). 
}

  \label{fig:precip_results}
\end{figure}

Figure \ref{fig:precip_results} presents a visual comparison between the ground truth precipitation maps and the model's corresponding predictions at three representative timestamps from 2020. Each column displays one timestamp, with the first row showing the IR input data, the second row showing the true observed precipitation, and the third and fourth rows showing the models' precipitation estimations. The model effectively captures the spatial distribution and intensity patterns of precipitation across a variety of atmospheric scenarios. In particular, it reconstructs core precipitation regions with good structural similarity, aligning with the low RMSE metrics reported in Section \ref{sec:precip_quant}. Some finer spatial features may be smoothed in the predictions, but the broader storm patterns and coverage areas are well-preserved.

These results demonstrate that the enhanced global context understanding provided by GAIA's MAE-DINO architecture translates effectively to downstream tasks, particularly in capturing complex atmospheric phenomena like precipitation patterns. The model is able to achieve qualitatively and quantitatively cohesive results on precipitation tasks, despite limited training data taken 15 years before the evaluation period.

\subsubsection{Atmospheric River Segmentation}

\subsubsection{Quantitative Analysis}

We evaluate GAIA's AR segmentation abilities relative to an MAE-only baseline using an identical attention-based decoder and report the following metrics at both patch- and pixel-level granularity: F1 score, precision, recall, and accuracy (Table \ref{tab:ar_metric_comparison}). GAIA consistently outperforms MAE-only across all metrics except for patch-level recall, where the two models perform equally well. Keeping the decoder the same across models attributes these gains to the more informative embeddings GAIA produces. Adding DINO to MAE encourages discriminative features that prioritize objects and regions within satellite data, whereas MAE alone optimizes for local pixel reconstruction. Because atmospheric rivers are large, filamentary structures, we hypothesize that GAIA’s representations better preserve global continuity and boundaries, allowing the decoder’s attention to focus on the AR backbone and suppress background clutter. This results in cleaner, more contiguous segmentations evidenced by improved performance, particularly at the patch level.

\begin{table}[!htb]
\centering
\renewcommand{\arraystretch}{1.3}
\begin{tabular}{l|cc}
\hline
\rowcolor{gray!10} \textbf{Metric} & \textbf{GAIA} & \textbf{MAE} \\
\hline
Patch-level F1 score                              & \textbf{0.58} & 0.52  \\
\rowcolor{gray!5} Patch-level precision & \textbf{0.50} & 0.42 \\
Patch-level recall                          & \textbf{0.68} & \textbf{0.68}  \\
\rowcolor{gray!5} Patch-level accuracy (\%) & \textbf{85} & 80 \\
\hline
Pixel-level F1 score & \textbf{0.47} & 0.40  \\
\rowcolor{gray!5} Pixel-level precision & \textbf{0.40} & 0.33 \\
Pixel-level recall & \textbf{0.58} & 0.52  \\
\rowcolor{gray!5} Pixel-level accuracy (\%) & \textbf{91} & 89 \\
\hline
\end{tabular}
\captionsetup{skip=8pt}  
\caption{Performance comparison of GAIA and MAE-only AR segmentation models across patch- and pixel-level metrics. Best values per row are highlighted in bold. We threshold separately for each model based on the best F1 score: 0.2 for both models for pixel-level metrics, and 0.09 / 0.07 for GAIA / MAE respectively for patch-level metrics.}
\label{tab:ar_metric_comparison}
\end{table}

\subsubsection{Qualitative Analysis}

\begin{figure}[ht]
  \centering
  \includegraphics[width=\textwidth]{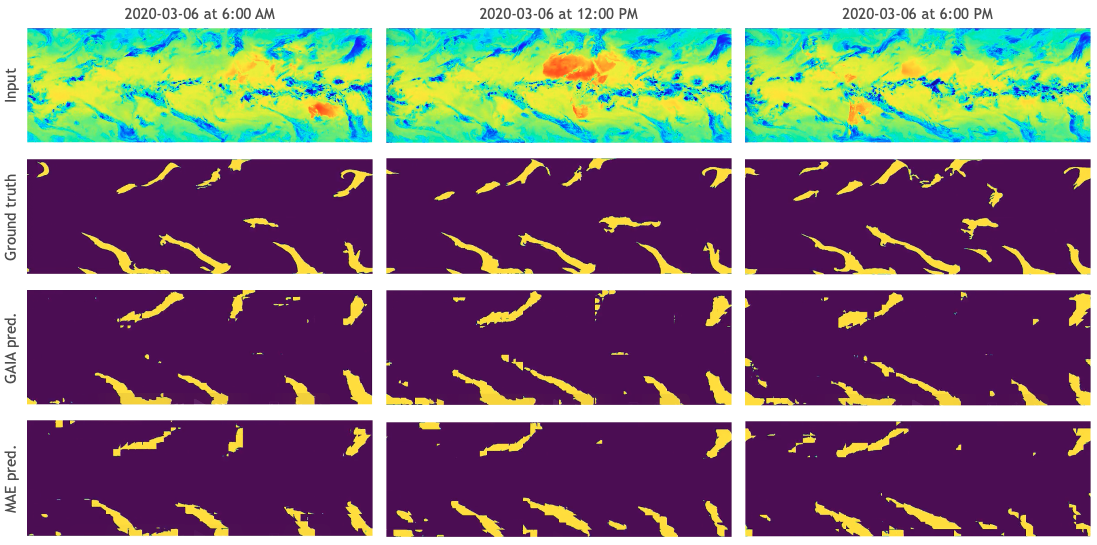}
  \caption{
Comparison of IR input (first row), ground truth (second row), and model-estimated AR segmentation (third and fourth rows) for three sequential timestamps in 2020 (left to right: 6 AM, 12 PM, 6 PM on March 6, 2020). Yellow reflects true (or predicted) atmospheric rivers while purple denotes background non-AR pixels.
}

  \label{fig:ar_results}
\end{figure}

Figure~\ref{fig:ar_results} illustrates three consecutive scenes: the top row shows the input satellite image, the second row the reference AR mask, and the bottom two rows GAIA’s and MAE's predicted segmentation, respectively. These figures reveal that GAIA can effectively group AR pixels together across patch boundaries, whereas the MAE predictions demonstrate many more patch-aligned segmentation boundaries (as evidenced by the jagged, axis-aligned borders of the segmentation mask). Additionally, the connectedness of each individual atmospheric river in the GAIA prediction indicates strong semantic understanding of each AR as a distinct object,  yielding clean contiguous masks when visualized. By contrast, the MAE baseline segmentation exhibits less confident boundaries, evidenced by small, isolated regions of positive predictions surrounding the principal atmospheric river prediction. The model additionally displays strong temporal consistency despite the lack of temporal information during training, demonstrating a strong understanding of AR temporal dynamics despite inference on each timestep in isolation. These behaviors reflect the quantitative gains in Table~\ref{tab:ar_metric_comparison} and are consistent with the encoder’s DINO component encouraging object-centric, augmentation-invariant features that align patch tokens with coherent regions rather than local textures. 

We also note characteristic failure modes that point to straightforward improvements in future research. Faint blockiness is visible along some regions, reflecting the large encoder patch size ($30 \times 30$); future iterations with smaller patch sizes could both improve pixel-level metrics and sharpen segmentation masks. Additionally, the model occasionally outputs AR detections that mirror baroclinic cloud bands observable in the input image, which suppresses model precision. This could indicate a limitation with the IR data we used, which may be insufficient for distinguishing cloudy non-ARs from ARs. 

Overall, qualitative assessment of our model — contiguous masks, clean boundaries, and restrained background responses — supports our claim that adding DINO to MAE produces more object-centric and separable patch embeddings suitable for detecting atmospheric rivers, enabling the same attention-based decoder to produce improved and cleaner AR segmentations with better generalization.

\subsubsection{Tropical Cyclone Detection}
To evaluate the GAIA architecture for TC detection, we compare its performance as a Mask-RCNN backbone against both an MAE-only and a supervised ResNet-FPN backbone's performance. All models are trained end-to-end to maximize performance on the detection task. We report the following performance statistics:

\begin{itemize}
    \item[(i)] \textbf{Frame-level detection metrics}: Box F1 score, precision, and recall at IoU $\geq$ 0.30 \footnote{We leverage a lower IoU threshold to mitigate impact of small, noisy data (i.e. uncentered bounding boxes).}.
    \item[(ii)] \textbf{Storm-level recall}: Fraction of known tropical cyclones (TCs) with at least one frame containing a valid detection.\footnote{We refer to valid detections as detection with an IoU of 0.3 with the ground-truth.}
    \item[(iii)] \textbf{Early detection frequency}: Fraction of TCs that have at least one valid detection at or before the first ground-truth label time
\end{itemize}


\subsubsection{Quantitative Analysis}

Across benchmarks, GAIA outperforms the MAE-only model, while the supervised ResNet remains strongest overall (Table \ref{tab:tc_detection_metrics}). For all models, the optimal F1 decision threshold occurs at a 0.0 confidence threshold, indicating that prioritizing recall yields the best F1 in a per-frame evaluation scheme; this is consistent with clear foreground / background separation learned in the proposal and classification stages. \footnote{Because frames are evaluated independently every 30 minutes without temporal smoothing, detections that precede the label time are counted as false positives at the frame level, which depresses precision; we therefore also report early-detection metrics that explicitly value lead time.}

\begin{table}[!htb]
\centering
\renewcommand{\arraystretch}{1.3}
\begin{tabular}{l|cc|c}
\hline
\rowcolor{gray!10} \textbf{Metric} & \textbf{GAIA} & \textbf{MAE} & \textbf{ResNet-FPN} \\
\hline
F1                              & \textbf{0.28} & 0.24 & 0.34  \\
\rowcolor{gray!5} Precision     & \textbf{0.29} & 0.26 & 0.36 \\
Recall                          & \textbf{0.27} & 0.22 & 0.33 \\
\rowcolor{gray!5} Storm-Level Recall & \textbf{81\%}  & 75\%  & 90\% \\
Early Detection Frequency    & \textbf{29\%}  & 17\%  & 31\% \\
\hline
\end{tabular}
\vspace{0.5em}
\caption{Tropical cyclone detection performance comparison across backbones, computed over the month of October (cyclone-heavy) from 2019-2021. Metrics are bolded with respect to a comparison between MAE and GAIA.}
\label{tab:tc_detection_metrics}
\end{table}

The GAIA backbone enables the detector to localize a larger fraction of storms while maintaining higher precision and recall at a frame-level than MAE-only, a theme persistent across confidence thresholds. GAIA also improves early detection relative to dataset label time ($29\%$ early detection frequency vs.\ $17\%$). We hypothesize that these gains largely arise from the DINO self-distillation objective, which improves sub-patch spatial awareness, making our patch embeddings better suited for capturing the smaller TC cores in GOES imagery. Despite this increased spatial awareness, GAIA is still limited in the detection task by its large (\(30\times30\)) patch-size, which creates an inherently coarse feature representation that prevents us from outperforming the supervised ResNet-FPN approach. This is a structural disadvantage of ViT-style backbones in comparison to ResNet-FPN, which benefits from real multi-resolution features. Future research will be dedicated to investigating the performance of the GAIA backbone for TC detection at lower patch sizes, which we suspect will close the gap between GAIA and ResNet.



\subsubsection{Qualitative Analysis}

\begin{figure}[h!]
  \centering
  \begin{subfigure}[b]{0.48\linewidth}
    \centering
    \includegraphics[width=\linewidth]{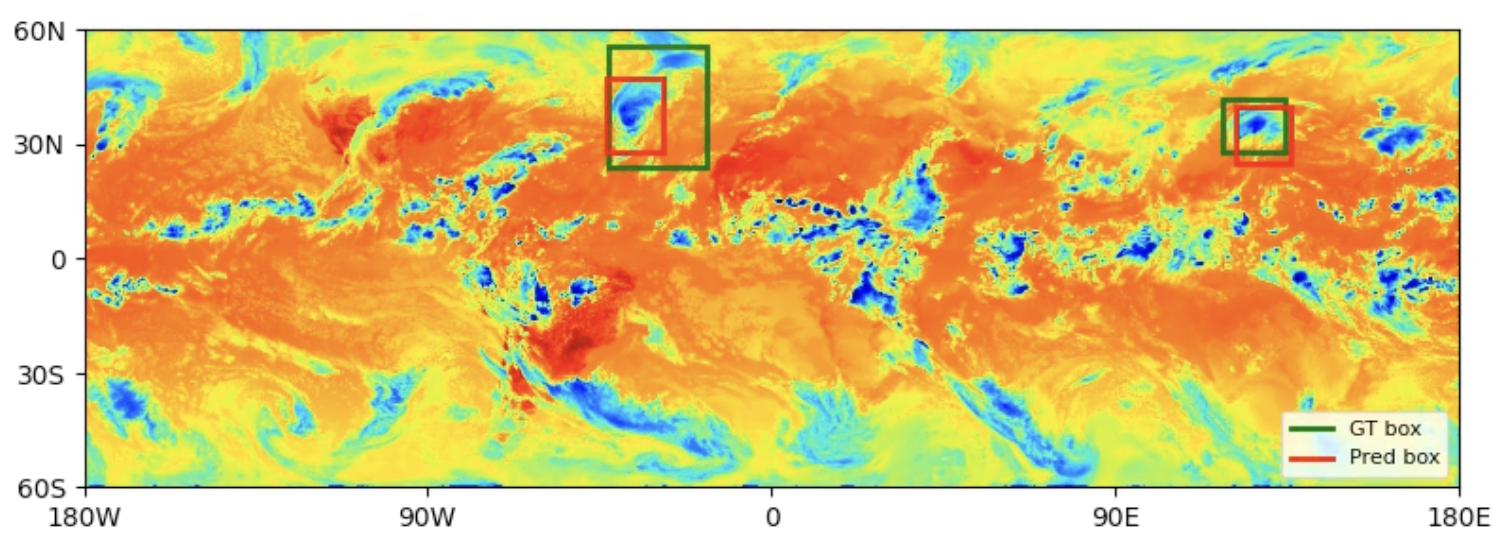}
    \caption{Predicted BBoxes + GT [2019-10-01 18:00]}
  \end{subfigure}\hfill
  \begin{subfigure}[b]{0.48\linewidth}
    \centering
    \includegraphics[width=\linewidth]{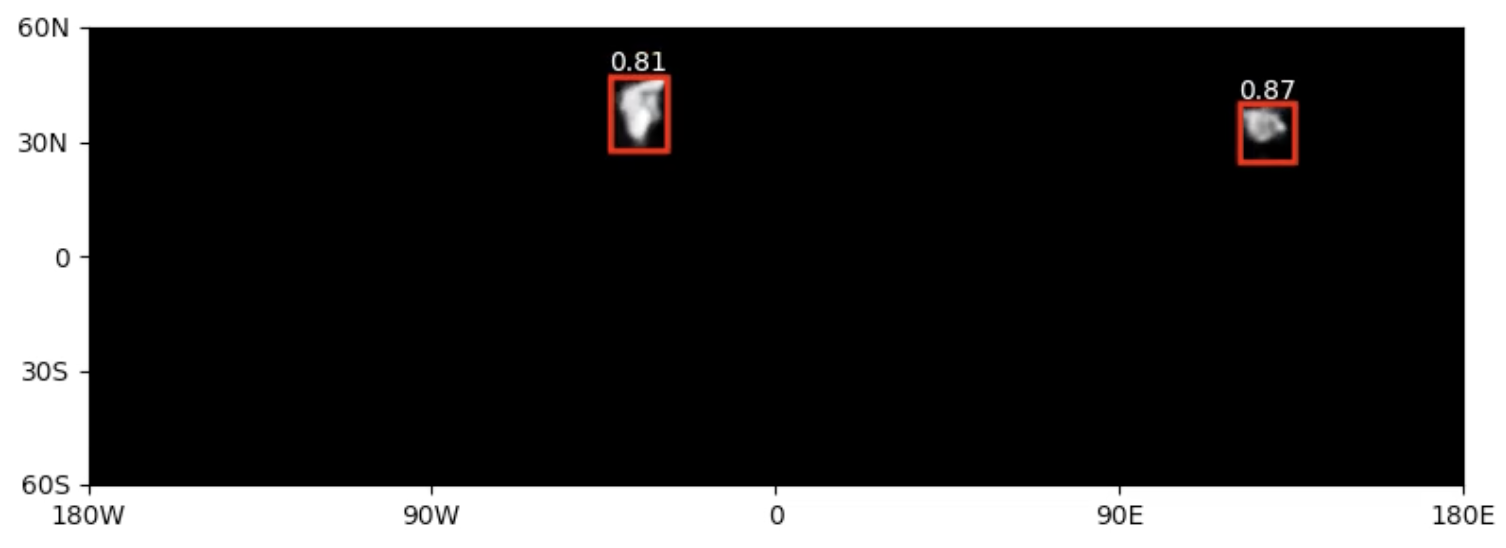}
    \caption{Predicted Masks [2019-10-01 18:00]}
  \end{subfigure}

  \vspace{0.5em}

  \begin{subfigure}[b]{0.48\linewidth}
    \centering
    \includegraphics[width=\linewidth]{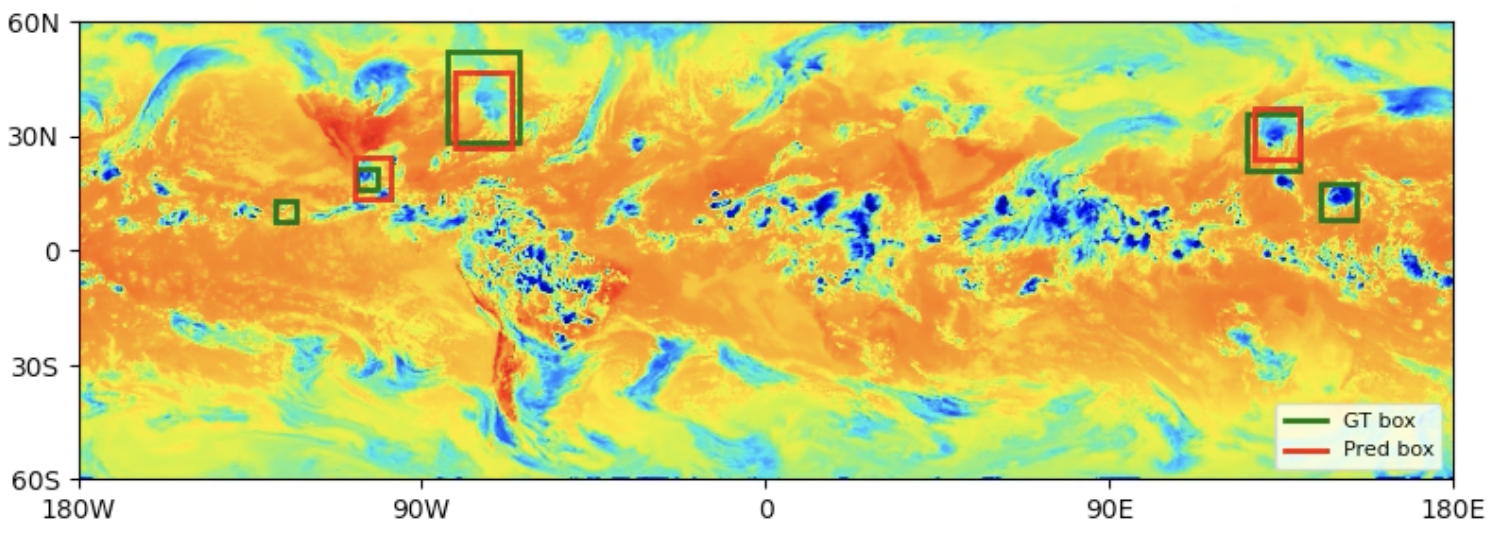}
    \caption{Predicted BBoxes + GT [2019-10-20 19:00]}
  \end{subfigure}\hfill
  \begin{subfigure}[b]{0.48\linewidth}
    \centering
    \includegraphics[width=\linewidth]{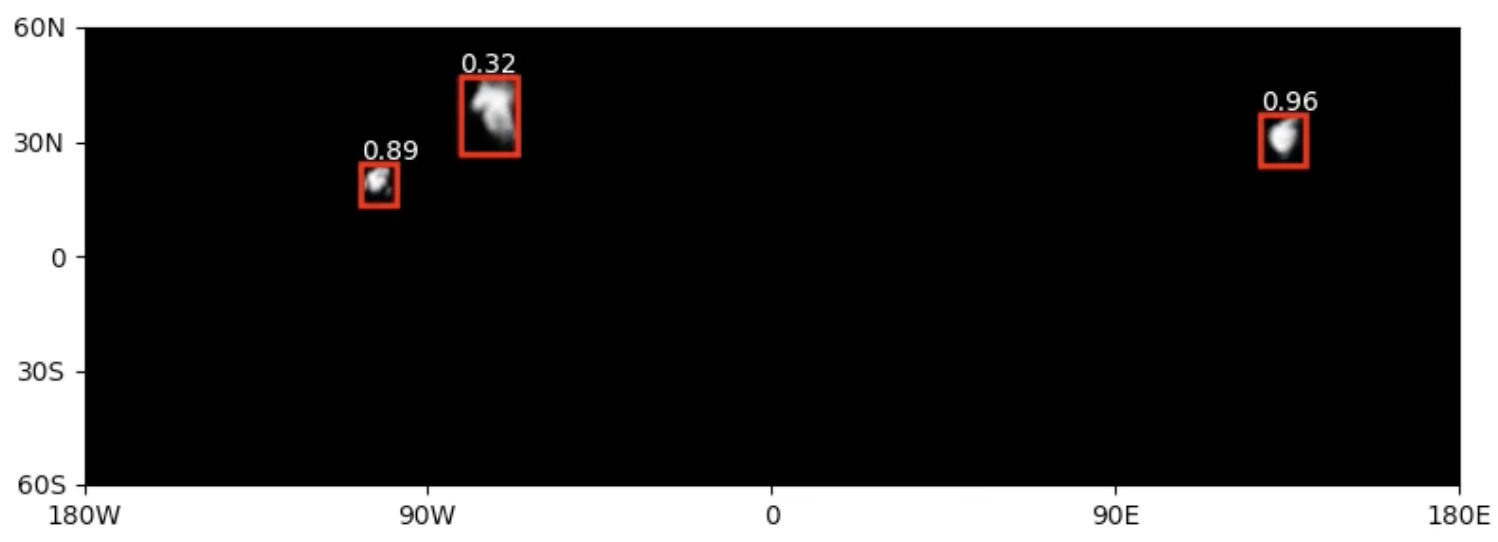}
    \caption{Predicted Masks [2019-10-20 19:00]}
  \end{subfigure}

  \vspace{0.5em}

  \begin{subfigure}[b]{0.48\linewidth}
    \centering
    \includegraphics[width=\linewidth]{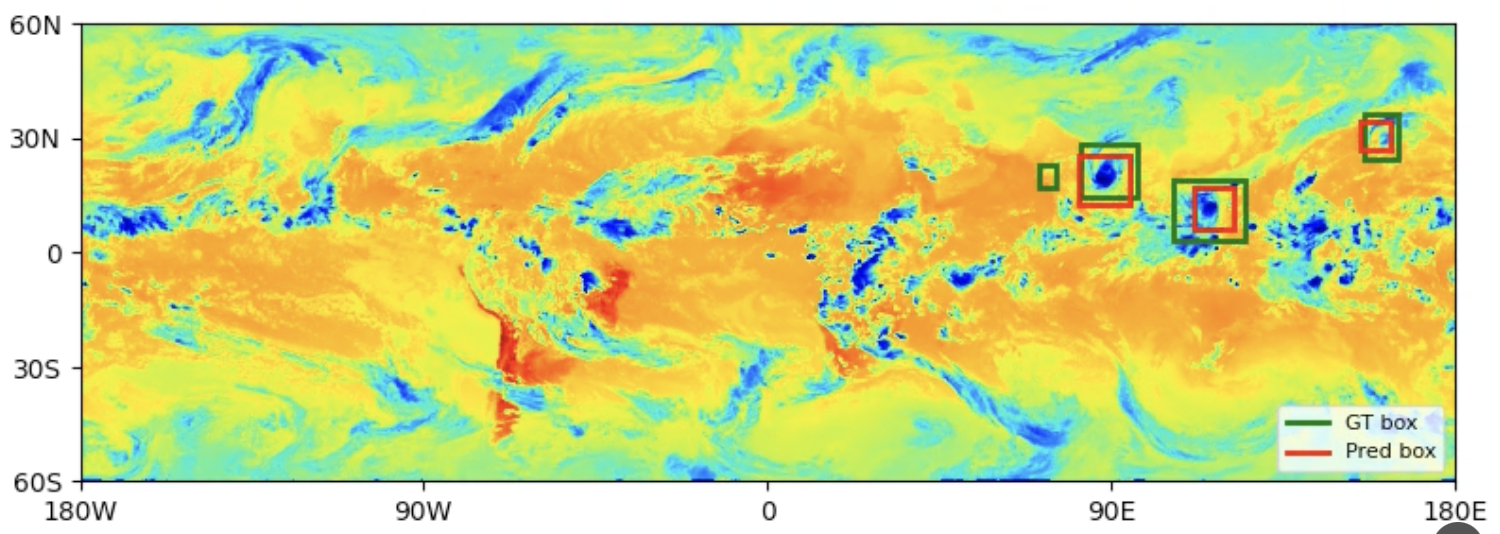}
    \caption{Predicted BBoxes + GT [2019-11-08 16:00]}
  \end{subfigure}\hfill
  \begin{subfigure}[b]{0.48\linewidth}
    \centering
    \includegraphics[width=\linewidth]{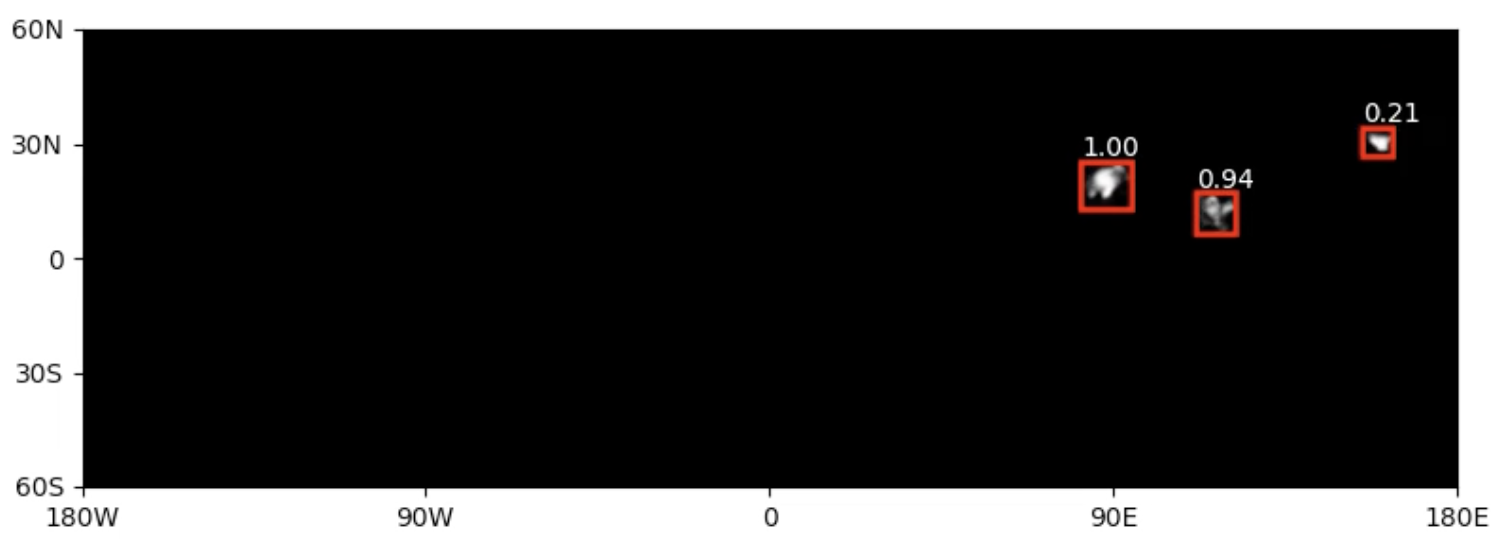}
    \caption{Predicted Masks [2019-11-08 16:00]}

  \end{subfigure}

  \caption{GAIA-backbone Mask-RCNN predicted boxes and segmentation masks. Predicted bounding boxes outlined in red, ground truth (GT) boxes highlighted in green.}
  \label{fig:maskrcnn-grid}
\end{figure}

Figure~\ref{fig:maskrcnn-grid} shows three representative GOES frames with predictions from Mask-RCNN using GAIA as the backbone. In all scenes, the detector correctly localizes multiple tropical cyclones that differ in size and are separated by large geographic distances, illustrating that the model is location-agnostic and can resolve several systems in a single pass. The left panels highlight the close alignment between predicted boxes and ground-truth annotations, and the right panels show that segmentation masks concentrate on cyclone rotational cores. Our detector largely ignores background thunderstorms and anvil clouds along the equator, which is consistent with the precision improvements reported in Table~\ref{tab:tc_detection_metrics}. The confidence scores annotated next to detections span from marginal ($\sim$0.2) to very high ($\sim$0.8–0.99). Operating at a low score threshold therefore surfaces weak or developing systems without materially increasing clutter, mirroring the higher storm-level recall achieved by GAIA. Finally, the ability to detect both small, compact cores and larger, more diffuse systems in the same frame highlights the benefit of GAIA features within the feature pyramid, aligning with the quantitative gains observed in Table \ref{tab:tc_detection_metrics}.

\begin{figure}[h!]
  \centering
  \begin{subfigure}{0.48\textwidth}
    \includegraphics[width=\linewidth]{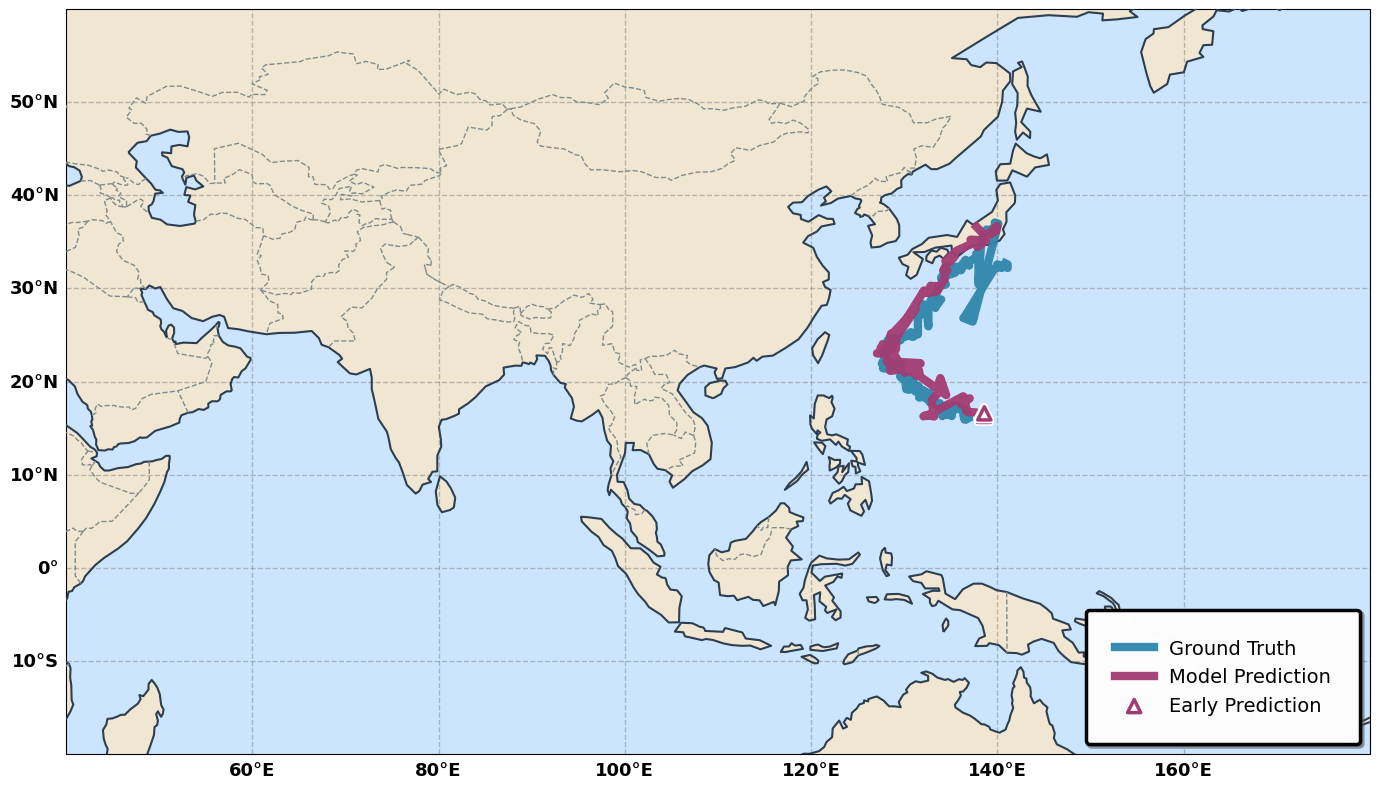}
    \caption{Hurricane Neoguri concatenated trajectory}
    \label{fig:neoguri}
  \end{subfigure}\hfill
  \begin{subfigure}{0.48\textwidth}
    \includegraphics[width=\linewidth]{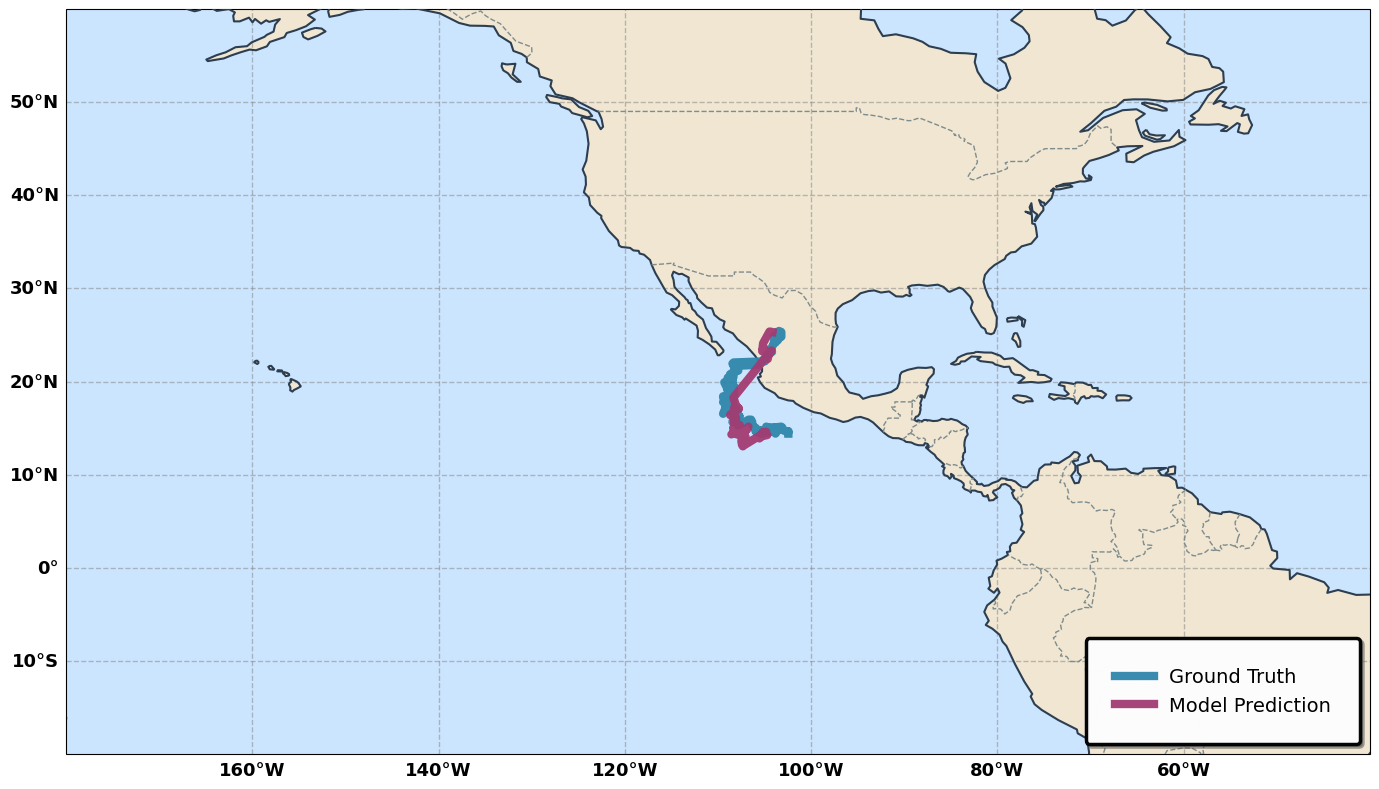}
    \caption{Hurricane Pamela concatenated trajectory}
    \label{fig:pamela}
  \end{subfigure}\hfill
  \caption{TC concatenated model trajectories vs. ground truth. Model displays strong temporal consistency through ability to track storms through independent frame-level prediction. Model trajectories (purple), ground truth (blue)}
  \label{fig:trajectories}
\end{figure}

Additionally, Figure \ref{fig:trajectories} highlights the temporal consistency of our model by concatenating individual predictions across timesteps, formulating a cyclone trajectory. The presence of temporal consistency despite the lack of temporal features represents an important benefit of the model, as it showcases our model's ability to understand cyclone evolution over time. Furthermore, sub-figure \ref{fig:neoguri}, which displays the predictions over time for Hurricane Neoguri, showcases the model's ability to detect storm formation early, as the storm is flagged 6 hours prior to being available in our ground-truth data.

\section{Conclusion}

We present GAIA, a novel hybrid self-supervised foundation model for atmospheric satellite imagery analysis that achieves superior representations by combining pre-training objectives—MAE and DINO—relative to a single-objective approach. Trained on 15 years of global geostationary observations, GAIA learns semantically rich embeddings that capture genuine atmospheric dynamics while exhibiting robust transfer capabilities across diverse operational meteorology tasks. Our comprehensive evaluation establishes both theoretical insights into representation learning and practical improvements for critical applications in weather monitoring and climate science.

A central contribution of our work lies in demonstrating that GAIA's dual pre-training objectives produce superior representations that avoid the trivial solutions often learned by single-objective models. Through principal component analysis, we show that GAIA learns semantically organized embeddings with distributed variance across multiple components (72.5\% cumulative across PC1-3)—a signature of rich, disentangled representations. In stark contrast, MAE- and DINO-only models exhibit highly concentrated variance (>88\%) where a single principal component dominates, indicating encodings that primarily capture low-frequency spatial information and positional structure. Visual analysis validates this interpretation: GAIA embeddings reveal distinct, spatially coherent features aligned with meaningful atmospheric zones—equatorial cloud belts, longitudinal circulation patterns—while baseline models produce embeddings that merely replicate smoothed versions of the input. Beyond spatial structure, our temporal analysis reveals that GAIA captures genuine atmospheric dynamics rather than memorizing diurnal solar patterns, with temporally adjacent observations clustering naturally without explicit temporal supervision. This spatiotemporal embedding structure proves critical for transfer learning, enabling different downstream tasks to leverage complementary subspaces of the learned representations.


GAIA exhibits strong intrinsic reconstruction capabilities across a wide range of mask ratios, maintaining low error from 30-50\% masking and graceful degradation even at extreme 90-95\% masking. This robustness extends to real-world gap filling applications, where GAIA consistently outperforms MAE-only baselines across synthetic and authentic missing data patterns, achieving superior SSIM and RMSE metrics. The model's ability to handle extreme masking during inference—despite pre-training at 75\%—demonstrates that learned representations generalize beyond specific training conditions.

The ultimate validation of GAIA's superior representations comes from consistent improvements across three operationally critical atmospheric science tasks. For precipitation estimation, the model maintains competitive performance despite limited training data. More compellingly, for atmospheric river segmentation—a task requiring coherent object-level understanding—GAIA achieves substantial gains over MAE-only baselines (F1: 0.58 vs. 0.52, patch-level accuracy: 85\% vs. 80\%), with qualitative analysis revealing notably more coherent, object-centric segmentation masks that respect atmospheric river boundaries across patch borders. For tropical cyclone detection, GAIA demonstrates clear advantages across multiple operational metrics: 81\% storm-level recall versus 75\% for MAE-only, and critically, 29\% early detection frequency versus only 17\%, representing a 70\% relative improvement in early warning capability. These consistent gains across diverse task types—dense regression, semantic segmentation, instance detection—provide strong empirical validation of our core hypothesis: combining complementary MAE and DINO objectives produces more flexible, transferable embeddings than either approach in isolation.

While our results are promising, several areas warrant future investigation. First, our relatively large patch size (30×30 pixels) and low-resolution input inherently limits fine-scale detail capture, particularly for compact structures like tropical cyclone cores. Future work should explore smaller patch sizes and native resolution GOES data to improve pixel-level performance and enable detection of finer atmospheric features. Second, incorporating multi-modal data—such as microwave observations or numerical weather prediction outputs—could enhance physical consistency and improve performance on rare events underrepresented in observational data. Third, extending the framework to explicitly model temporal dependencies (e.g., through video prediction objectives) could improve forecasting capabilities. Finally, applying GAIA to other remote sensing domains, such as ocean dynamics or land surface monitoring, could validate the generalizability of the hybrid MAE-DINO approach beyond atmospheric applications.

GAIA's capabilities have significant implications for operational meteorology and climate science. The model's robust gap filling enables continuous monitoring despite sensor failures or data transmission errors, supporting reliable climate trend analysis. Early tropical cyclone detection could improve disaster preparedness and evacuation planning. Improved atmospheric river segmentation aids in flood forecasting for water resource management. More broadly, GAIA's transfer learning efficiency suggests the potential for rapid adaptation to new sensors, geographic regions, or atmospheric variables with limited labeled data—a critical capability as observational networks expand globally.

GAIA represents a significant step forward in bridging self-supervised learning methodology with operational atmospheric science. By rigorously demonstrating that hybrid pre-training objectives produce more semantic, disentangled, and transferable representations than single-objective approaches—evidenced through both interpretability analyses and consistent downstream performance gains—we contribute insights valuable beyond atmospheric applications to the broader foundation model community. The model's combination of theoretical understanding (avoiding trivial solutions, learning genuine dynamics) and practical utility (robust gap filling, improved early warning) establishes GAIA as a versatile foundation for addressing critical challenges in weather monitoring, climate analysis, and extreme event prediction. We release our pre-trained model weights and complete codebase openly to accelerate both scientific research and operational deployment in the atmospheric science community.

\newpage
\section*{Acknowledgments}
This work used resources available through the USRA Research Institute for Advanced Computer Science (RIACS) and the National Research Platform (NRP) at the University of California, San Diego. NRP has been developed, and is supported in part, by funding from National Science Foundation, from awards 1730158, 1540112, 1541349, 1826967, 2112167, 2100237, and 2120019, as well as additional funding from community partners. The author(s) would like to thank Alexander Lobo, Evan Muscatel, An Nguyen, Sam Solovy, Claudia Jimenez Arellano, Besart Mujeci, and Dima Mishin for their invaluable help and insightful discussions throughout the development of this work.

\bibliographystyle{unsrt}
\bibliography{references}



\end{document}